\pdfoutput=1
\documentclass{egpubl}
\usepackage{egsr2021}
 
\SpecialIssuePaper         %

\usepackage[T1]{fontenc}
\usepackage{dfadobe}  

\biberVersion
\BibtexOrBiblatex
\usepackage[backend=biber,bibstyle=EG,citestyle=alphabetic,backref=true, natbib=true]{biblatex} 
\electronicVersion
\PrintedOrElectronic

\ifpdf \usepackage[pdftex]{graphicx} \pdfcompresslevel=9
\else \usepackage[dvips]{graphicx} \fi

\usepackage{egweblnk}

\usepackage{booktabs} %
\usepackage{xcolor}
\usepackage{colortbl}
\usepackage{xspace}
\usepackage{csquotes}
\usepackage{subcaption}
\usepackage{multirow}
\usepackage{microtype}
\usepackage[colorinlistoftodos,prependcaption,textsize=small]{todonotes}
\usepackage[acronym]{glossaries}
\usepackage[binary-units = true, range-phrase=--, load=prefixed]{siunitx}
\DeclareSIUnit\flops{FLOPS}

\setglossarysection{paragraph}
\makeglossaries

\newcommand{\revised}[1]{{#1}}

\newcommand{\ourtodo}[1]{}
\newcommand{\ourtodoinline}[1]{}
\newcommand{\ourtodolow}[1]{}
\newcommand{\ours}{\gls{donerf}\xspace}

\newacronym{nerf}{NeRF}{neural radiance field}
\newacronym{hmd}{HMD}{head-mounted display}
\newacronym{mlp}{MLP}{multilayer perceptron}
\newacronym{nsr}{NSR}{neural scene representation}
\newacronym{vr}{VR}{virtual reality}
\newacronym{ar}{AR}{augmented reality}
\newacronym{ibr}{IBR}{image-based rendering}
\newacronym{mpi}{MPI}{multi-plane image}
\newacronym{nsvf}{NSVF}{neural sparse voxel field}
\newacronym{srn}{SRN}{scene representation network}
\newacronym{nv}{NV}{neural volume}
\newacronym{derf}{DeRF}{decomposed radiance field}
\newacronym{llff}{LLFF}{Local Light Field Fusion}
\newacronym{nex}{NeX}{Neural Basis Expansion}
\newacronym{donerf}{DONeRF}{depth oracle neural radiance field}

\glsdisablehyper

\newcommand{\eg}{\emph{e.g.}\xspace}
\newcommand{\ie}{\emph{i.e.}\xspace}

\newcommand{\etal}{et~al.\xspace}
\newcommand{\bulldozer}{\emph{Bulldozer}\xspace}
\newcommand{\forest}{\emph{Forest}\xspace}
\newcommand{\classroom}{\emph{Classroom}\xspace}
\newcommand{\pavillon}{\emph{Pavillon}\xspace}

\newcommand{\sanmiguel}{\emph{San Miguel}\xspace}
\newcommand{\barbershop}{\emph{Barbershop}\xspace}

\let\vec\mathbf

\usepackage[ruled]{algorithm2e} %

\SetAlFnt{\small}
\SetAlCapFnt{\small}
\SetAlCapNameFnt{\small}
\SetAlCapHSkip{0pt}

\title[DONeRF: Towards Real-Time Rendering of Compact Neural Radiance Fields using Depth Oracle Networks]%
      {DONeRF: Towards Real-Time Rendering of Compact Neural Radiance Fields using Depth Oracle Networks}

\author[Neff et al.]
{\parbox{\textwidth}{\centering T. Neff$^{1}$\orcid{0000-0002-6559-5653}, P. Stadlbauer$^{1}$\orcid{0000-0003-1199-9641}, M. Parger$^{1}$\orcid{0000-0002-9074-4374}, A. Kurz$^{1}$\orcid{0000-0001-9151-1006}, J. H. Mueller$^{1}$\orcid{0000-0002-6368-6340}, C. R. A. Chaitanya$^{2}$, A. Kaplanyan$^{2}$\orcid{0000-0002-8376-6719} and M. Steinberger$^{1}$\orcid{0000-0001-5977-8536}
}
\\
{\parbox{\textwidth}{\centering $^1$Graz University of Technology, Austria\\
         $^2$Facebook Reality Labs, USA
}
}
}

\begin{document}

\teaser{
 \centering
	\captionsetup{labelfont=bf,textfont=it,justification=centering}
	\begin{subfigure}[t]{.163\linewidth}
		\includegraphics[width=\linewidth]{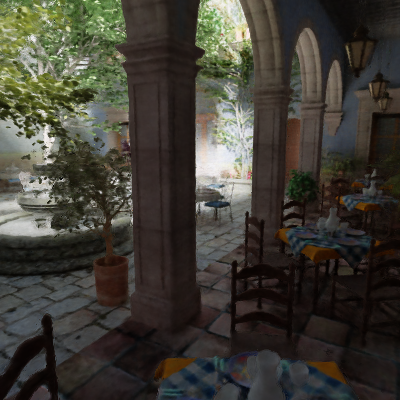}
		\caption{\sanmiguel \\ (\revised{$27.41$dB})}
	\end{subfigure}
	\begin{subfigure}[t]{.163\linewidth}
		\includegraphics[width=\linewidth]{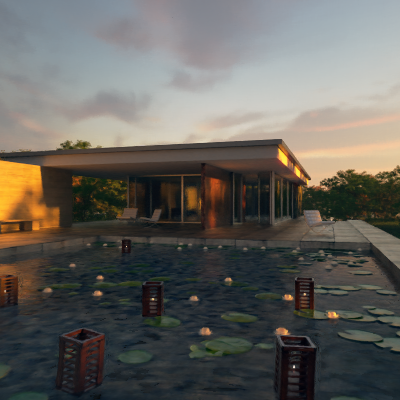}
		\caption{\pavillon \\ (\revised{$31.07$dB})}
	\end{subfigure}
	\begin{subfigure}[t]{.163\linewidth}
		\includegraphics[width=\linewidth]{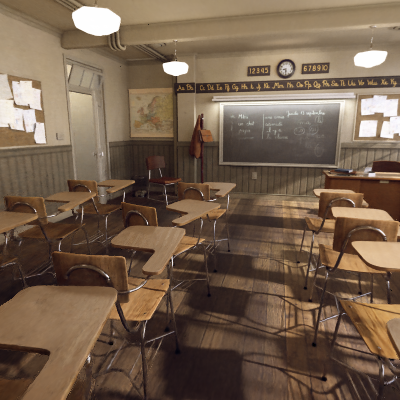}
		\caption{\classroom \\ (\revised{$33.43$dB})}
	\end{subfigure}
	\begin{subfigure}[t]{.163\linewidth}
		\includegraphics[width=\linewidth]{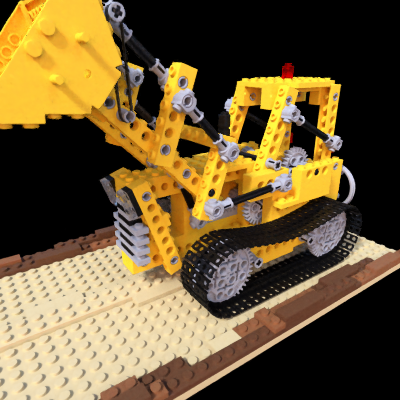}
		\caption{\bulldozer \\ (\revised{$33.46$dB})}
	\end{subfigure}
	\begin{subfigure}[t]{.163\linewidth}
		\includegraphics[width=\linewidth]{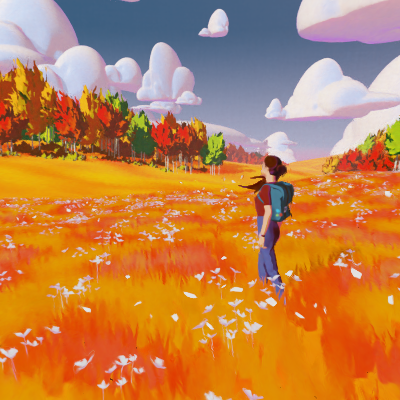}
		\caption{\forest \\ (\revised{$30.63$dB})}
	\end{subfigure}
	\begin{subfigure}[t]{.163\linewidth}
		\includegraphics[width=\linewidth]{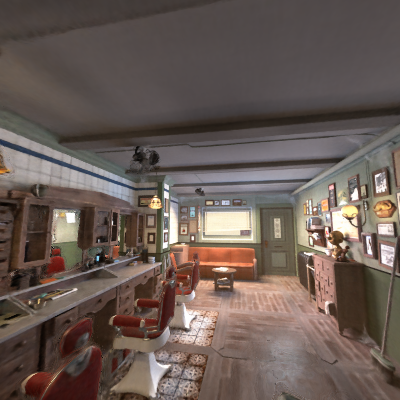}
		\caption{\barbershop \\ (\revised{$30.84$dB})}
	\end{subfigure}
	\captionsetup{justification=justified}
	\caption{Example renderings with DONeRF at $400\times400$ pixels for our tested scenes (PSNR is in brackets). All shown results are rendered in real-time at \SI{22}{\ms} per frame on a single GPU and require approximately $4.35$ MFLOP per pixel to compute. DONeRF requires only 4 samples per pixel thanks to a depth oracle network to guide sample placement, while NeRF uses 256 samples per pixel in total. We reduce the execution and training time by up to $48\times$ and achieve better quality (NeRF average PSNR at \revised{$30.52$} dB vs. our \revised{$31.14$} dB).}
	\label{fig:teaser}
}

\maketitle
\begin{abstract}
The recent research explosion around implicit neural representations, such as NeRF, shows that there is immense potential for implicitly storing high-quality scene and lighting information in compact neural networks.
However, one major limitation preventing the use of NeRF in real-time rendering applications is the prohibitive computational cost of excessive network evaluations along each view ray, requiring dozens of petaFLOPS.
In this work, we bring compact neural representations closer to practical rendering of synthetic content in real-time applications, such as games and virtual reality.
We show that the number of samples required for each view ray can be significantly reduced when samples are placed around surfaces in the scene without compromising image quality.
To this end, we propose a depth oracle network that predicts ray sample locations for each view ray with a single network evaluation.
We show that using a classification network around logarithmically discretized and spherically warped depth values is essential to encode surface locations rather than directly estimating depth.
The combination of these techniques leads to DONeRF, our compact dual network design with a depth oracle network as its first step and a locally sampled shading network for ray accumulation.
With DONeRF, we reduce the inference costs by up to 48x compared to NeRF when conditioning on available ground truth depth information.
Compared to concurrent acceleration methods for raymarching-based neural representations, DONeRF does not require additional memory for explicit caching or acceleration structures, and can render interactively (20 frames per second) on a single GPU.


\end{abstract}

\section{Introduction}

Real-time rendering of photorealistic scenes with complex lighting is still an overly demanding problem. %
However, today, consumer machine learning accelerators are widespread from desktop GPUs to mobile phones and \gls{vr} headsets, making evaluation of neural networks fast and power-efficient.
Recent advances in implicit neural scene representations~\cite{sitzmann2019scene,nn2019siren,mildenhall2020nerf} %
impressively show that machine learning can be used for compact encoding and high-quality rendering of 3D scenes.
\Glspl{nerf}~\cite{mildenhall2020nerf} use only \num{1000000} parameters divided among two \gls{mlp} networks to encode scene structure alongside lighting effects.
For image generation, \gls{nerf} uses traditional volume rendering drawing \num{256} samples for each view ray, where each sample requires a full network evaluation.

Although \gls{nerf}-like methods show significant potential for compact high-quality object and scene representations, they are too expensive to evaluate in real-time.
Real-time rendering of such a representation onto a \gls{vr} headset at $1440\times 1600$ pixel per eye with \SI{90}{\Hz} would require \num{37} petaFLOPS (\num{256} network evaluations each with $256^2\cdot7$ multiply add operations).
Clearly, this is not possible on current GPU hardware and evaluation cost is a major limiting factor for neural representations to be used for real-time rendering.
Additionally, \gls{nerf} only works well for small scale content, requiring splitting larger scenes into multiple \glspl{nerf} \cite{zhang2020nerf, rebain2020derf}, multiplying both the memory and evaluation cost.

In this work, we make neural representations practical for interactive and real-time rendering, while sticking to a tight memory budget.
Particularly, our goal is to enable large scale \emph{synthetic content} in movie quality in real-time rendering.
We make the following contributions with \glspl{donerf}:
\begin{itemize}
	\item We propose a compact dual network design to reduce evaluation costs for neural rendering. An \emph{oracle network} predicts sample locations along view rays and a \emph{shading network} places a small number of samples guided by the oracle to deliver the final color.
	\item We present a robust \emph{depth oracle network} design and training scheme to efficiently provide sample locations for the shading network.
	The oracle uses filtered, discretized target depth values, which are readily available in synthetic content, and it learns to solve a classification task rather than to directly estimate depth.
	\item We introduce a non-linear transformation to handle large, open scenes and show that sampling of the shading network should happen in a warped space, to better capture different frequencies in the fore- and background, capturing content in a single network beyond the capability of previous work.
	\item Combining our efforts, we demonstrate high-quality real-time neural rendering of large synthetic scenes. %
	At the same tight memory budget used by the original \gls{nerf}, we show equal quality for small scenes and significant improvements for large scenes, while reducing the computational cost by \numrange{24}{98}$\times$.
\end{itemize}

With \ours, we are the first to render large-scale computer graphics scenes from a compact neural representation in real time.
Additionally, \ours is significantly faster to train. %
We focus on static synthetic scenes and consider dynamic scenes and animations orthogonal to our work.
Still, \ours can directly be used as a compact backdrop for distant parts of a game scene, or in \gls{vr} and \gls{ar}, where an environment map does not offer the required parallax for a stereo stimulus.
Our source code and datasets are available at \url{https://depthoraclenerf.github.io/}.

\begin{figure*}[!ht]
	\centering
	\captionsetup{labelfont=bf,textfont=it}
	\includegraphics[width=.78\linewidth]{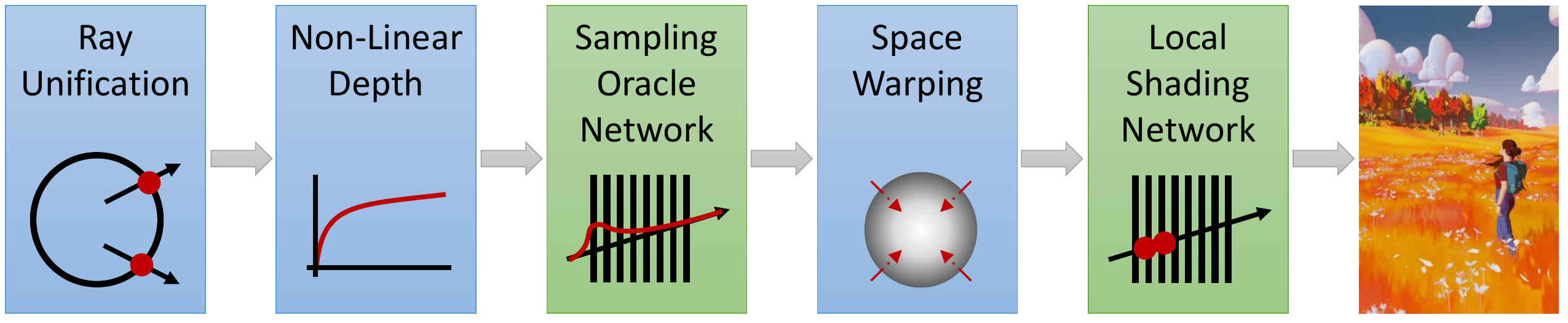}
	\caption{To enable efficient rendering of large-scale neural representations, \ours uses a five stage pipeline: (1) ray descriptions are unified within a view cell, (2) depth is considered in a non-linear space, (3) an oracle network estimates the importance of samples, (4) sample positions are warped towards the view cell, and (5) radiance is generated from the shading \gls{mlp} with only a few samples along each ray.}
	\label{fig:overview}
\end{figure*}

\section{Related work}
\paragraph*{Image-based novel view synthesis}
Recently, image-based rendering techniques using \glspl{mpi}~\cite{zhou2018_mpi, flynn_2019_deepview} managed to achieve impressive results by blending image layers. %
By blending between multiple \glspl{mpi}~\cite{mildenhall2019llff} and using spherical image layers~\cite{broxton2020immersive} the potential field of view can be increased at the cost of memory efficiency.
Further extending \glspl{mpi}, neural basis functions can be learned to enable real-time view synthesis~\cite{Wizadwongsa2021NeX}. %
Alternatively, an implicit mapping between view, time or illumination conditions can be learned~\cite{bemana2020x}.
Although explicit image-based representations can be efficiently rendered, they typically only allow for small viewing regions~\cite{srinivasan_2019_mpi_boundaries} in addition to requiring densely sampled input images, which substantially increases memory requirements for larger viewing regions.

\paragraph*{Implicit neural scene representations}
Although explicit neural representations based on voxels~\cite{sitzmann2019deepvoxels}, \glspl{mpi}~\cite{zhou2018_mpi, flynn_2019_deepview} or proxy geometry~\cite{DeepBlending2018} enable fast novel view generation, they are fundamentally limited by the internal resolution of their representation. %
To circumvent this issue, implicit neural scene representations~\cite{park2019deepsdf, sitzmann2019scene} directly infer outputs from a continuous input space, such as ray origins and directions.
\Glspl{srn}~\cite{sitzmann2019scene} directly map 3D world coordinates to a feature representation and use a learned raymarcher to accumulate rays for the final RGB output.
Similarly, neural volumes~\cite{Lombardi:2019} use raymarching to accumulate rays in a learned, warped, volumetric representation. %
The quality of scene representations can be improved with periodic activation functions~\cite{nn2019siren}.

\paragraph*{Neural Radiance Fields}
Opening a whole new subdomain of research, Mildenhall \etal \cite{mildenhall2020nerf} introduced \gls{nerf}, which replaced the learned raymarching from \gls{srn} with a fixed, differentiable ray marcher. %
In \gls{nerf}, all ray samples are transformed into a high dimensional sine-cosine or Fourier space~\cite{tancik2020fourfeat}, and fed into an \gls{mlp}, followed by an accumulation step to generate the final RGB output. 
The simplicity and impressive results inspired many adaptations to the original \gls{nerf}, sometimes being referred to as the \emph{\gls{nerf} explosion}~\cite{dellaert2021neural}:
\glspl{nerf} can capture dynamic free-viewpoint video~\cite{park2020nerfies, li2020neural, xian2020space, pumarola2020d, du2020nerflow}, generate photorealistic avatars~\cite{gafni2020dynamic, gao2020portraitnerf, lombardi2021mixture}, perform relighting on captured scenes~\cite{martinbrualla2020nerfw, bi2020neural, nerv2020, boss2020nerd}, conditionally encode shape and appearance via latent codes~\cite{Schwarz2020NEURIPS, chanmonteiro2020pi-GAN, yu2020pixelnerf, grf2020} and compose scenes of multiple objects~\cite{ost2020neural, yuan2021star, niemeyer2020giraffe}.

Although these \gls{nerf} variants show impressive quality, the large number of samples per ray typically makes \glspl{nerf} unsuitable for real-time applications.
As a result, several recent publications incrementally improve run-time efficiency. %
To enable empty space skipping,\glspl{nsvf}~\cite{liu2020neural} uses a self-pruning sparse voxel octree structure,  where each ray sample includes information from a tri-linearly interpolated embedding of voxel vertices.
Alternatively, to reduce the number of evaluations along a ray, partial integrals can be learned~\cite{lindell2020autoint}.
\Glspl{derf}~\cite{2020arXiv201112490R} decompose the scene with a Voronoi decomposition to train multiple \glspl{nerf} for each cell. %

\paragraph*{Baking of Neural Radiance Fields}
Recent research has focused on baking  components of \gls{nerf} to achieve performance gains at the cost of extensive memory consumption~\cite{yu2021plenoctrees, hedman2021snerg, reiser2021kilonerf, garbin2021fastnerf}.
While baking could also be applied to our work, 
it departs from the beauty of a compact neural representation, potentially requiring hundreds of \si{\mega\byte}s up to \si{\giga\byte}s for a scene that can be represented by \SI{4}{\mega\byte} in \gls{nerf} or our approach.

In our work, we increase the inference speed of \gls{nerf}-like representations while staying in the realm of compact \glspl{mlp} without additional data structures or increased storage requirements.
At the memory requirement of two \glspl{mlp}, we show the most significant performance improvements compared to \gls{nerf}~\cite{mildenhall2020nerf}. %
Additionally, we increase image quality and support large-scale scenes, where the original \gls{nerf}, image-based methods, and methods that require additional data structures, like \gls{nsvf}~\cite{liu2020neural}, struggle.

\section{Efficient Neural Rendering using Depth Oracle Networks}
\label{sec:method}

To achieve real-time rendering of compact neural representations for generated content, we introduce \ours.
\ours replaces the \gls{mlp}-based raymarching scheme of \gls{nerf}~\cite{mildenhall2020nerf} with  a compact local sampling strategy %
to only consider important samples around surfaces. 
\ours consists of two networks in a five-stage pipeline (Figure~\ref{fig:overview}): 
A \emph{sampling oracle network} predicts optimal sample locations along the view ray using classification and a \emph{shading network} uses \gls{nerf}-like raymarching accumulation to deliver RGBA output.
To remove input ambiguity, we transform rays to a \emph{unified space} and use \emph{non-linear sampling} to focus on close regions.
Between the two networks, we warp the local samples to direct high frequency predictions of the shading network to the foreground.

\begin{figure}
	\centering
	\captionsetup{labelfont=bf,textfont=it}
	\includegraphics[width=0.79\linewidth]{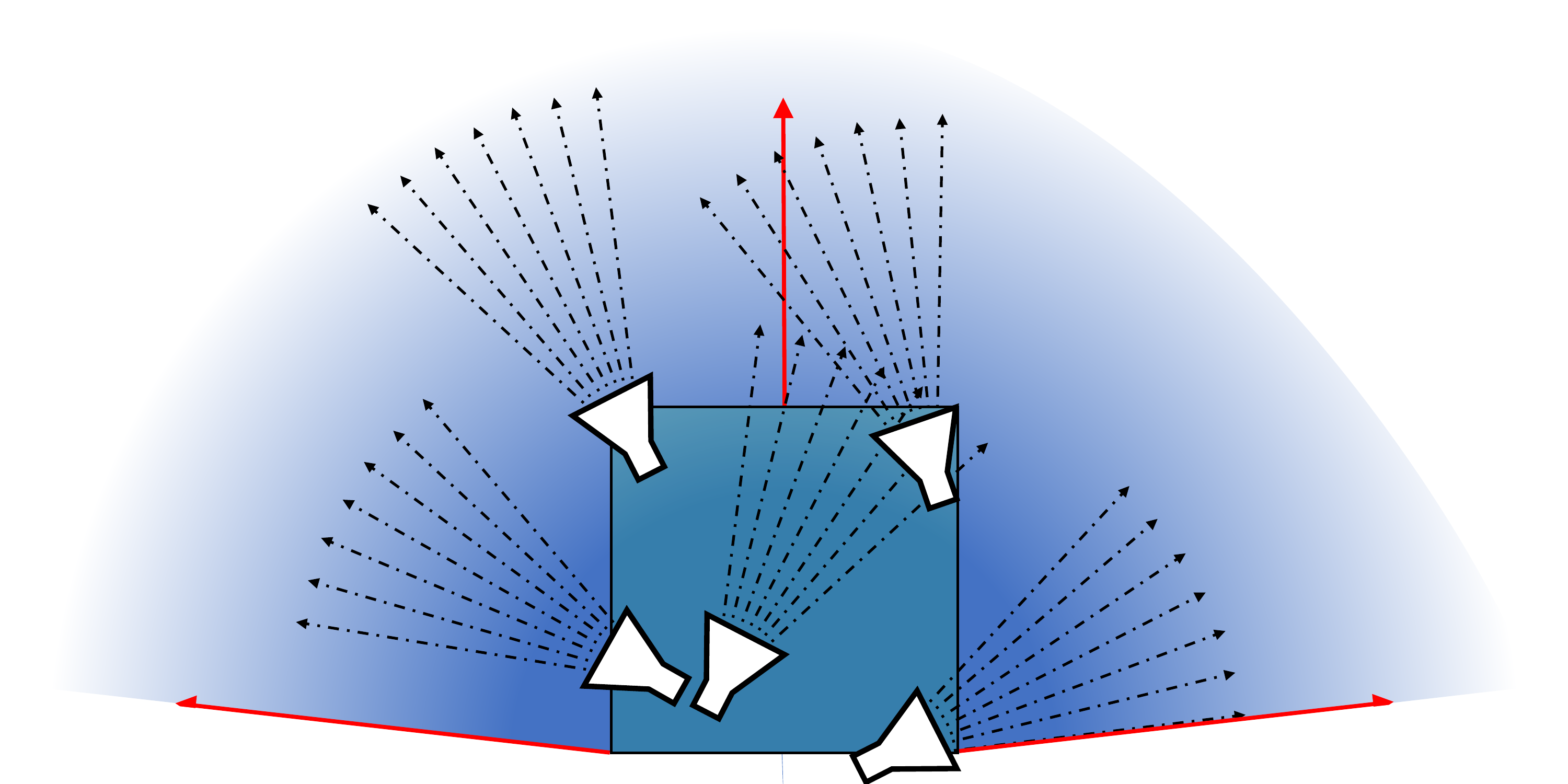}
	\caption{Top view of a view cell, defined by a bounding box, a forward direction (arrow) and a maximum viewing angle (blue with limiting arrows). Valid rays originate within the view cell and stay within the angle bounds (see example camera orientations).}
	\label{fig:viewcell}
\end{figure}

\paragraph*{View Cells}
For training, we use RGBD input images sampled from a \emph{view cell}.
A view cell is defined as a bounding box with a primary orientation and maximum viewing angle, \ie, it captures all view rays that originate in the bounding box and stay within a certain rotation, see Figure \ref{fig:viewcell}.
In a streaming setup, trained network weights for partially overlapping view cells can be swapped or interpolated to enable seamless transitions between larger sets of potential views.
\revised{We define the view cell specifics to provide a clear way of splitting large scenes and defining the potential input to our approach.
As smaller view cells reduce the visible content of a scene, smaller view cells may lead to higher quality (with the extreme being a single view).
However, large view cells can work similarly well while being even more memory efficient---depending on the network capacity used to represent them.
Note that this is true for any scene representation, and comes at a cost of larger memory requirements.
}

\section{Efficient Neural Sampling}
\label{sec:nerf_sampling}

Inference performance of \gls{nerf}-like neural representations scales most significantly with the number of samples per ray. 
While there are methods that deploy multiple lower-capacity networks for a moderate quality-speed tradeoff~\cite{rebain2020derf}, the target output quality is fundamentally limited by the given network capacity.
Thus, we consider network capacity optimizations orthogonal to our work.
Instead, we consider different sample placement strategies to reduce the amount of samples for neural raymarching.

\paragraph*{Uniform Sampling}
The default way of \gls{nerf}-style raymarching is to place samples uniformly between the near and far plane:
\begin{align}
	\vec x(d_i) & = \vec o + d_i  \cdot \vec r \\
	d_i & = \left ( d_{min} + i \cdot \frac{(d_{max} - d_{min}) }{N}\right ), \  i=[0, 1, 2, \cdots, N], \nonumber
\end{align}
\revised{where $\vec o$ is the ray origin, $\vec r$ is the ray direction, $N$ is the number of placed samples, and $d_{min}$ and $d_{max}$ are the near and far plane distances from the camera pose.}
Sample locations are transformed to a view cell local coordinate system, divided by $d_{max}$, and positionally encoded to construct the feature vector $\vec f$, where $\vec c$ is the view cell center:
\begin{equation}
	\vec f(d_i) = \text{encode}\left ( \frac{\vec x(d_i) - \vec c}{d_{max}}  \right ).
\end{equation}

\paragraph*{Non-linear Sampling}
\label{sec:nonlinear_sampling}
While uniform sampling works well for individual objects and small scenes, large depth ranges are problematic. %
Focusing samples on objects closer to the camera 
is an intuitive first step to reduce network evaluations without losing quality. 
For non-linear sample placement, we use a logarithmic non-linearity:
\begin{equation}
\tilde{d}_i = d_{min} + \frac{\log(d_i - d_{min} + 1)}{\log(d_{max} - d_{min} + 1)} \cdot (d_{max} - d_{min}).
\end{equation}

\revised{
\paragraph*{NDC Sampling}
\gls{nerf}~\cite{mildenhall2020nerf} suggests to transform rays into an average camera frame, and to uniformly sample within the projected normalized device coordinates (NDC), directly feeding NDC samples into positional encoding.
This approach is only applicable if all samples lie strictly in the front hemisphere of the average camera frame, which is a limitation not shared by the other sampling strategies.
Large deviations from the average camera frame lead to increasingly high non-linear perspective distortions, which may impact the learning process.
We refer the reader to the appendix in the original \gls{nerf} paper for the detailed derivations; we use the same transformation with $d_{min} = 1$ and $d_{max} = \infty$.
}

\paragraph*{Space Warping}
\label{sec:space_warping}
Although non-linear sampling focuses the samples on the foreground, positional encoding is applied equally for fore- and background. 
Early training for large scenes showed that the background often contains high frequencies that must be dampened by the network while the foreground requires those to achieve sufficient detail.
This is not surprising, as real cameras and graphics techniques such as mip-mapping average background details.

\begin{figure}
	\captionsetup{labelfont=bf,textfont=it}
	\begin{subfigure}[t]{.32\linewidth}
		\includegraphics[width=\linewidth,page=1]{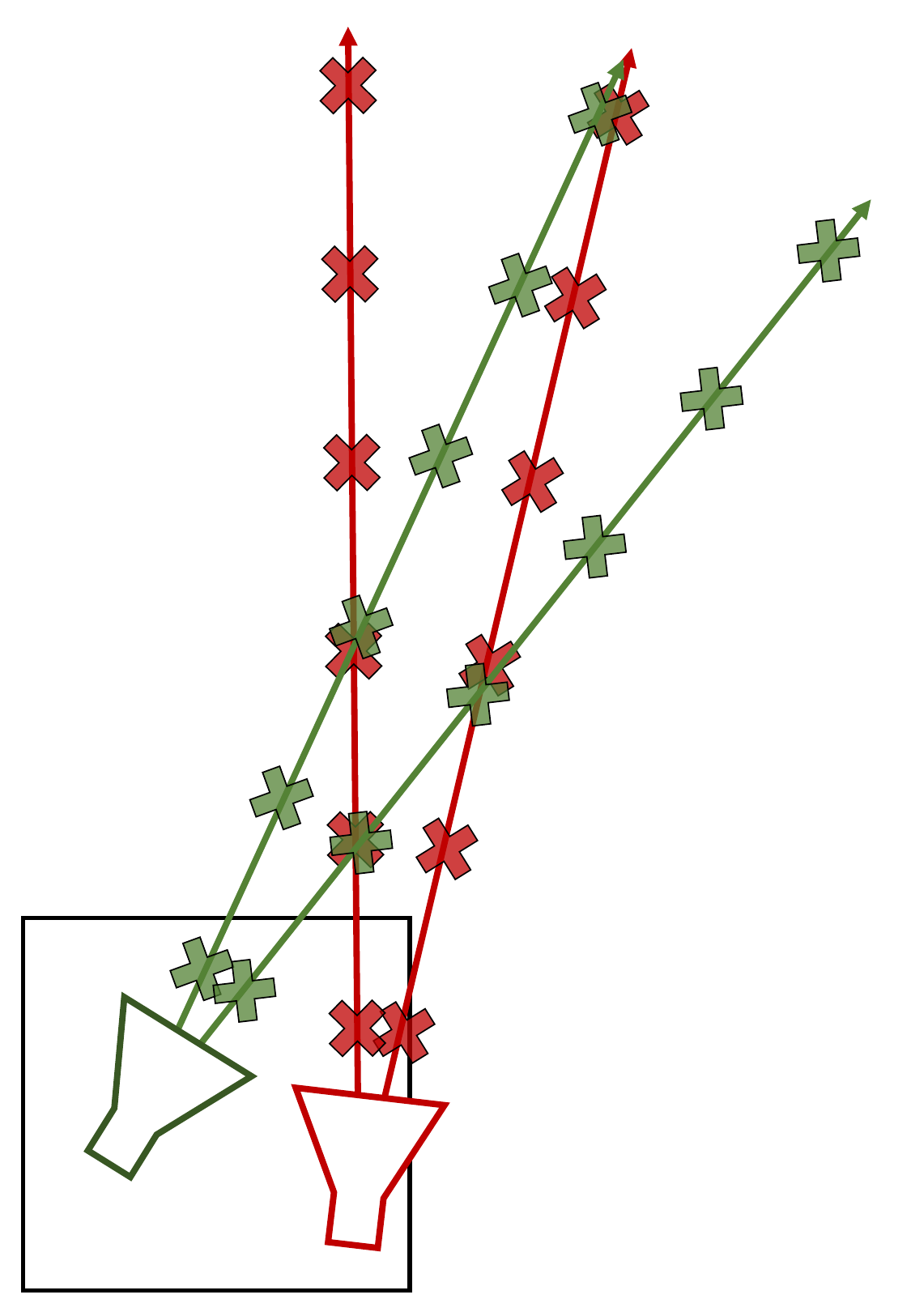}
		\caption{uniform}
	\end{subfigure}%
	\begin{subfigure}[t]{.32\linewidth}
		\includegraphics[width=\linewidth,page=2]{figures/samplingmethods.pdf}
		\caption{logarithmic}
	\end{subfigure}%
	\begin{subfigure}[t]{.32\linewidth}
		\includegraphics[width=\linewidth,page=3]{figures/samplingmethods.pdf}
		\caption{log+warp}
	\end{subfigure}
	\caption{Different sampling approaches visualized: (a) uniform samples in equal steps between the near and far planes; (b) logarithmic sampling reduces the sample distance for close samples in favor of spacing out far samples; (c) log+warp pulls the space closer to the view cell center, making the scene appear smaller to the \gls{nerf} and bending rays (compare to the gray straight lines).}
	\label{fig:samplingmethods}
\end{figure}

To remedy this issue, we propose a warping of the 3D space towards the view cell center.
We warp the entire space using a radial distortion, bringing the background closer to the view cell for positional encoding.
Initial experiments showed that using an inverse square root transform works well:
\begin{align}
\tilde{\vec f} & = \text{encode} \left ( (\vec x(\tilde{d}_i) - \vec c ) \cdot W(\vec x (\tilde{d}_i) - \vec c) \right ) \\
W(\vec p) & = \frac{1}{\sqrt{|\vec p|\cdot d_{max}}}.
\end{align}
While ray samples do not follow a straight line after warping, sample locations in space stay consistent, \ie, samples from different view points landing at the same 3D location are evaluated equally.
\revised{See Figure \ref{fig:samplingmethods} for a visualization of the \emph{uniform}, \emph{logarithmic} and \emph{log+warp} placement strategies.
The NDC sampling strategy is linear along each ray in NDC space, but follows a linear sampling in \emph{disparity} from the near plane to infinity in the original space, with a more aggressive $\frac{1}{x}$ sampling curve compared to logarithmic sampling. 
From the perspective of the network input, it therefore combines the uniform and logarithmic sampling approaches.
}

\paragraph*{Local Sampling}
\label{sec:localsampling}
Even when focusing samples on the foreground, %
 \glspl{nerf} with large sample counts spend many samples in empty space.
Given a ground truth surface representation, \eg, a depth texture, %
it is possible to take a fraction of the samples of a trained \gls{nerf} around the surface and still achieve mostly equal quality.
This inspires the following question: \emph{Given a ground truth depth texture to place samples during training, what is the best quality-speed tradeoff that can possibly be reached?}

\paragraph*{Ablation Study}

To determine the effectiveness of the different sampling strategies, we run an ablation study based on the original \gls{nerf}.
For all experiments, we assume static geometry and lighting and test on a single view cell. %
We vary the number of samples per ray $N$ between $[2, 4, 8, 16, 32, 64, 128]$ and only use a single \gls{mlp} (without the refinement network proposed in the original \gls{nerf} work which dynamically influences sample placement). 
To investigate local sampling, we perform \emph{uniform}, \emph{logarithmic}, \emph{log+warp} and \emph{NDC} sampling around the ground truth depth, keeping the step size identical to $N = 128$ for all sample counts.
We conduct our experiments on four \revised{diverse} scenes, \bulldozer, \forest, \classroom and \sanmiguel. %
For more details on the evaluation setup and the evaluated test scenes, please refer to Section~\ref{sec:evaluation} and Appendix~\ref{app:baselines}.

\begin{figure}
	\centering
	\captionsetup{labelfont=bf,textfont=it}
	\includegraphics[width=1.0\linewidth]{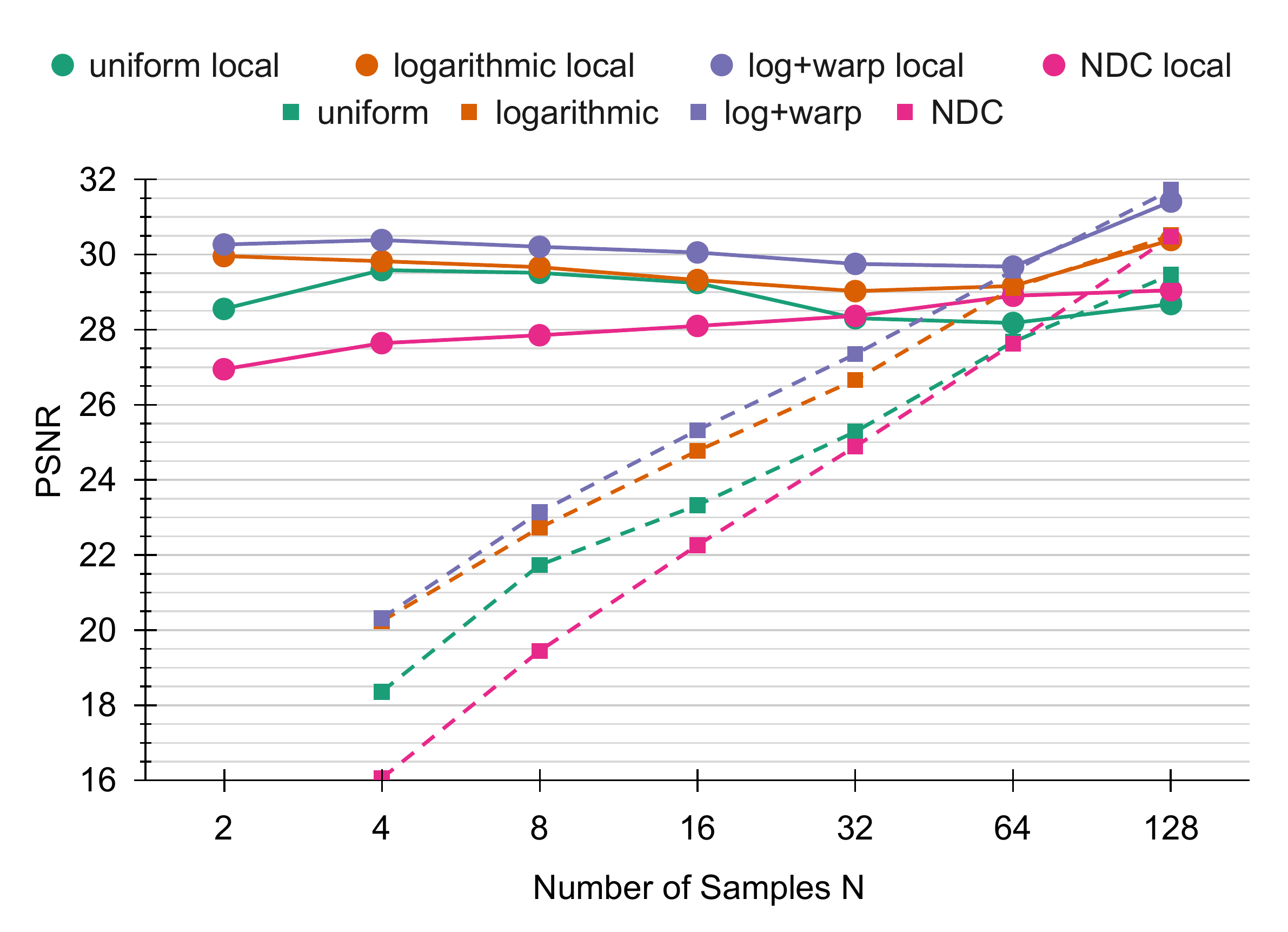}
	\caption{Average PSNR results for various sample reduction schemes over the sample count $N$. 
		\emph{Uniform} sampling roughly increases quality by $1.8$ dB for each doubling of $N$.
		\emph{logarithmic} increases PSNR by \revised{$1.3$ dB} and  inverse square root warping (\emph{log+warp}) adds another \revised{$0.6$ dB} on top.
		\revised{Sampling in \emph{NDC} only manages to surpass uniform sampling at $128$ samples.}
		When sampling around the ground truth depth (\emph{local}), the number of samples hardly affects quality. Still, PSNR roughly increases by $0.75$ dB with \emph{logarithmic} and another \revised{$0.6$ dB} with \emph{log+warp}.
		\revised{Local \emph{NDC} sampling performs worse than all other local sampling approaches until $N=32$.}
	}
	\label{fig:nerf_num_samples_gt_depth_plot}
\end{figure}

Averaged results are shown in Figure \ref{fig:nerf_num_samples_gt_depth_plot}, while per-scene details %
 can be found in Appendix~\ref{app:nerf_sampling}.
First, for non-local sampling, the results show that \emph{logarithmic} sampling increases quality over \emph{uniform} while \emph{log+warp} increases quality further.
With \emph{log+warp} the number of samples can be halved to still achieve equal quality to \emph{uniform}.
Second, with local sampling, the number of samples can be reduced to two with nearly no decrease in quality.
On average, \emph{log+warp} adds about \revised{\SI{1.3}{\decibel}} in quality over \emph{uniform} for local sampling.
\revised{Sampling in \emph{NDC} only reaches competitive results at more than $64$ samples, falling behind \emph{uniform} sampling in our evaluation. Since the network inputs in NDC are still uniform, the underlying non-linear transformation between different rays must be reconstructed by the network, which seems difficult at lower sample counts.}

Our results indicate that---given an ideal sampling oracle---significant gains in quality-speed tradeoffs can be achieved \revised{by placing samples locally}.
However, in practice, ground truth depth is often not available during inference for neural rendering due to memory or computational constraints: large-scale scenes require a significant amount of storage to represent via geometry or individual depth maps, and reprojection would be necessary to generalize to novel views.
Thus local sampling from ground truth depth during inference can be considered a niche scenario, and we therefore target an efficient and compact representation via an \gls{mlp}-based sampling oracle that only uses ground truth depth during training. %

\section{Sampling Oracle Network}
\label{sec:sampling_oracle_network}

\begin{figure}
	\centering
	\captionsetup{labelfont=bf,textfont=it}
	\begin{subfigure}[t]{.32\linewidth}
		\includegraphics[width=\linewidth]{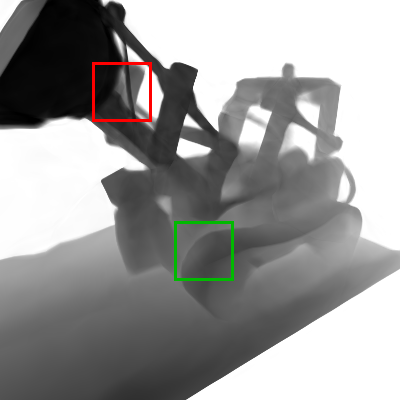}\\%
		\vspace{1pt}%
		\includegraphics[width=0.48\linewidth]{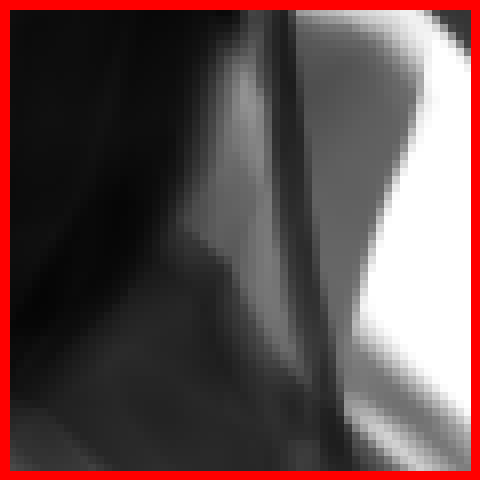}%
		\hfill%
		\includegraphics[width=0.48\linewidth]{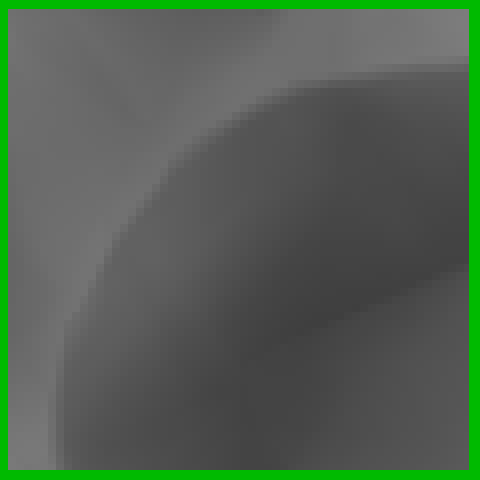}
		\caption{Depth Prediction}
	\end{subfigure}%
	\hfill%
	\begin{subfigure}[t]{.32\linewidth}
		\includegraphics[width=\linewidth]{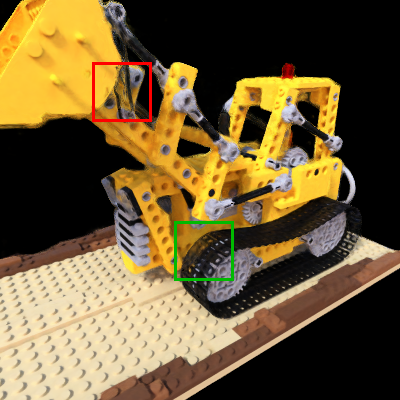}\\%
		\vspace{1pt}%
		\includegraphics[width=0.48\linewidth]{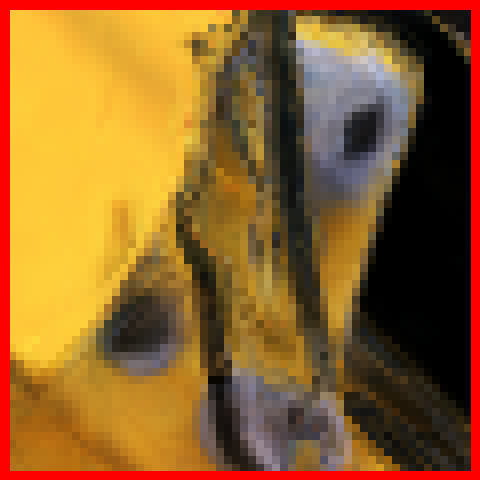}%
		\hfill%
		\includegraphics[width=0.48\linewidth]{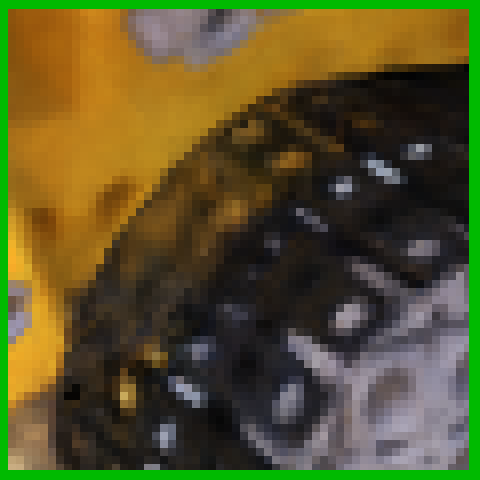}
		\caption{3 Local Samples}
	\end{subfigure}%
	\hfill%
	\begin{subfigure}[t]{.32\linewidth}
		\includegraphics[width=\linewidth]{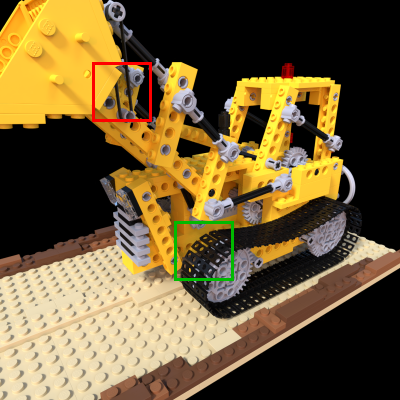}\\%
		\vspace{1pt}%
		\includegraphics[width=0.48\linewidth]{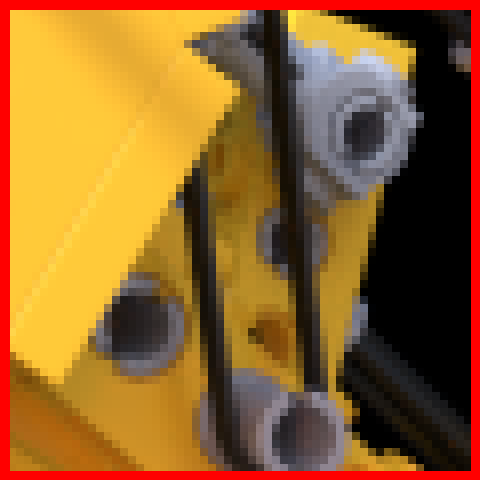}%
		\hfill%
		\includegraphics[width=0.48\linewidth]{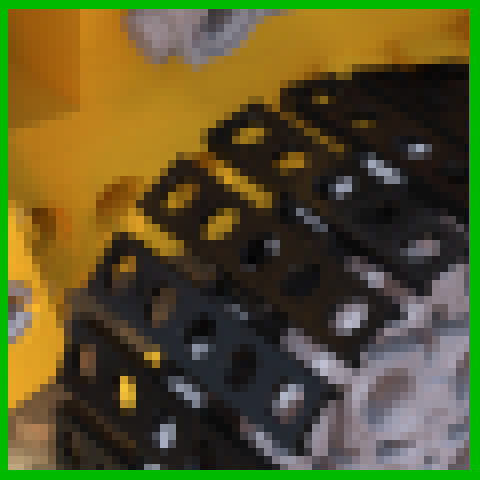}
		\caption{Ground Truth}
	\end{subfigure}%
	\caption{(a) A depth oracle network with a single depth output smooths depth around discontinuities. (b) As a result, local sampling mixes fore- and background and distorts features. (c) This becomes apparent when compared to the ground truth.}
	\label{fig:depthoutline}
\end{figure}

As mentioned before, sampling  around a known ground truth surface representation can reduce the number of required samples by up to $64\times$.
However, relying on explicit surface representations would defeat the purpose of having a compact neural representation.
Therefore, we introduce an oracle network to predict ideal sample locations for the raymarched shading network.
This oracle network takes a ray as input and provides information about sample locations along that ray.
We found that using an \gls{mlp} of the same size as the shading network generates sufficiently consistent depth estimates for the majority of rays in simple scenes.
However, accurate estimates around depth discontinuities remain difficult to predict, leading to significant visual artifacts (Figure \ref{fig:depthoutline}).

To mitigate this issue, we start with the following observation: In general, the exact geometric structure of the scene is faithfully reconstructed using neural raymarching, where samples are also placed in empty space.
Thus, we can allow the oracle more freedom. 
While it must predict sample locations around the surface, it may provide \emph{additional} estimates.
Consider neighboring rays around a depth discontinuity, either hitting the foreground or the background:
Representing this discontinuity accurately in ray space is difficult, as small changes in ray origin or direction may alternate between fore- and background.
However, if the oracle is allowed to return the same result for all rays around discontinuities, \ie, sampling at the fore- \emph{and} the background, the oracle's task is easier to learn.

\subsection{Classified Depth}
\label{sec:classified_depth}
Interestingly, a simultaneous prediction of multiple real-valued depth outputs did not improve results compared to a single depth estimate per pixel in our experiments.
Alternatively, the oracle can output sample likelihoods along the ray---\ie, a likelihood that a sample at a certain location will increase image quality.
Unlike \gls{nerf}'s refinement network, we only want to evaluate the oracle network once and %
reduce the sample count of the shading network. %

\begin{figure}
	\centering
	\captionsetup{labelfont=bf,textfont=it}
	\includegraphics[page=7,width=0.8\linewidth]{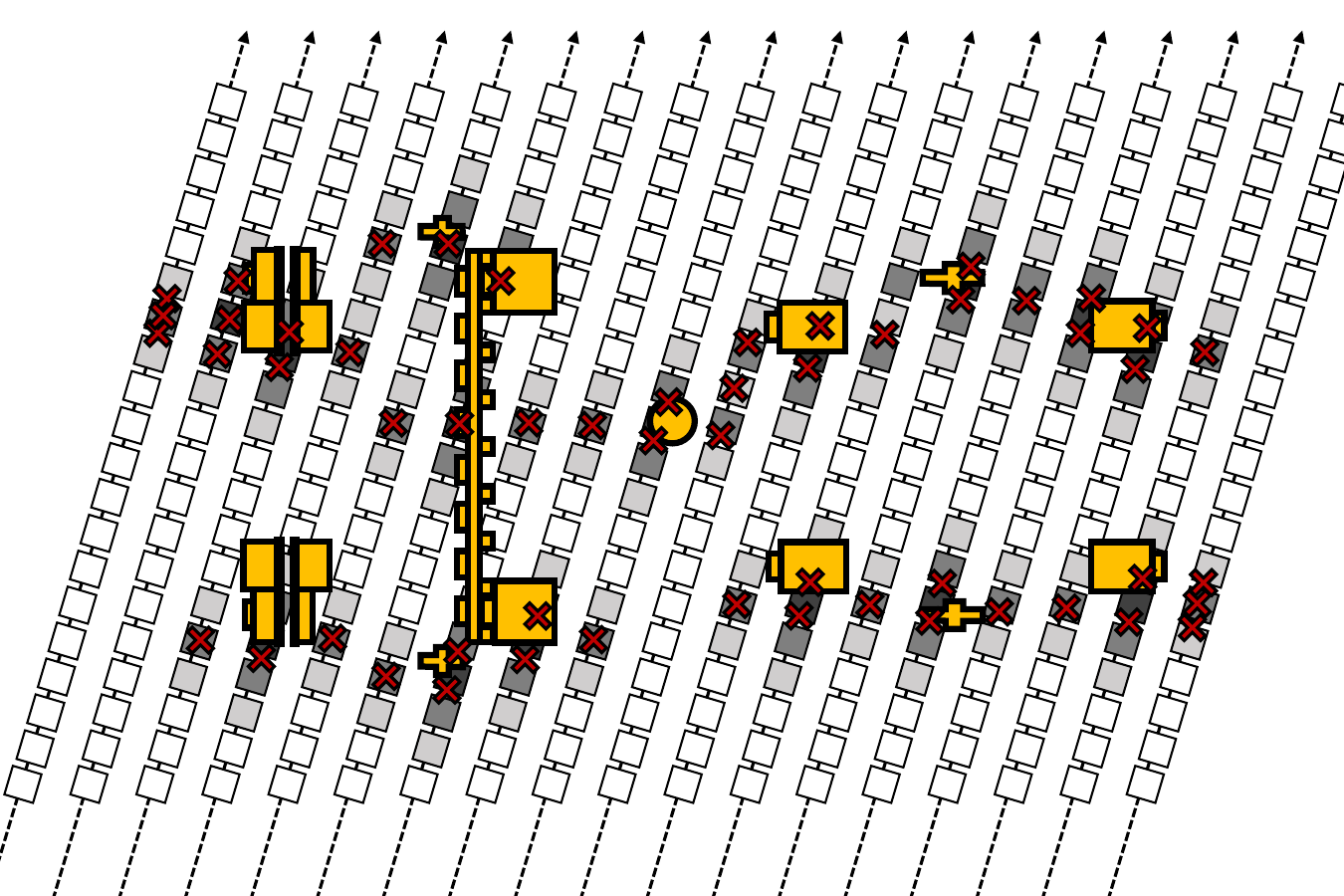}
	\caption{A horizontal slice through the \bulldozer dataset.
		Filtering the depth classification target (black = 1, white = 0) in both image dimensions and along depth smooths the classification target. An oracle producing such an output still results in high quality sample locations (red $\times$), with additional samples placed in free space. A smooth target is easier to learn as labels vary with lower frequency.
	}
	\label{fig:classified_depth_filtering_drawing}
\end{figure}

\begin{figure*}[t]
	\captionsetup{labelfont=bf,textfont=it}
	\begin{subfigure}[t]{.16\linewidth}
		\includegraphics[width=\linewidth]{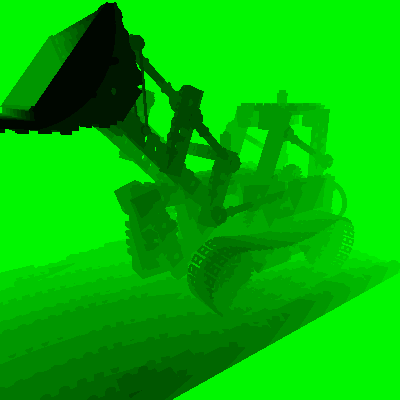}
		\caption{\bulldozer $K=1$}
	\end{subfigure}
	\begin{subfigure}[t]{.16\linewidth}
		\includegraphics[width=\linewidth]{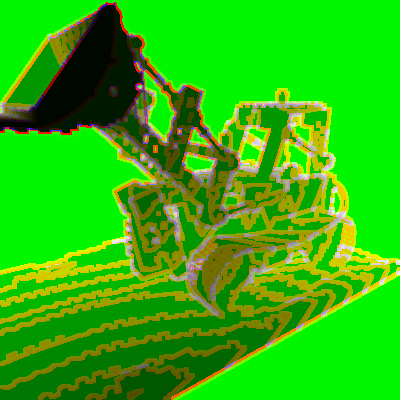}
		\caption{\bulldozer $K=5$}
	\end{subfigure}
	\begin{subfigure}[t]{.16\linewidth}
		\includegraphics[width=\linewidth]{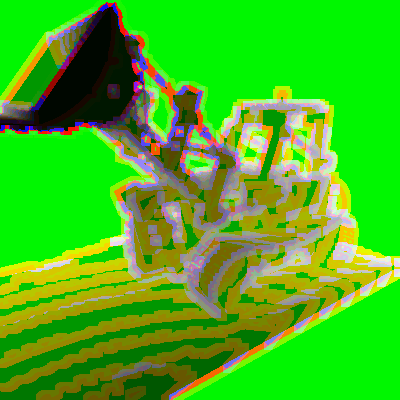}
		\caption{\bulldozer $K=9$}
	\end{subfigure}
	\begin{subfigure}[t]{.16\linewidth}
		\includegraphics[width=\linewidth]{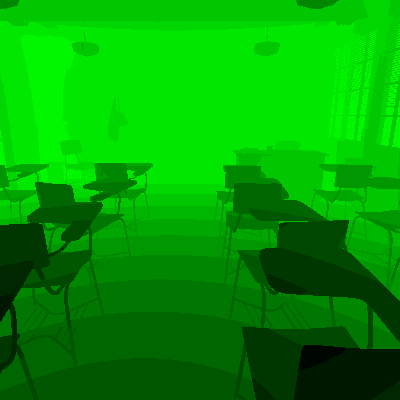}
		\caption{\classroom $K=1$}
	\end{subfigure}
	\begin{subfigure}[t]{.16\linewidth}
		\includegraphics[width=\linewidth]{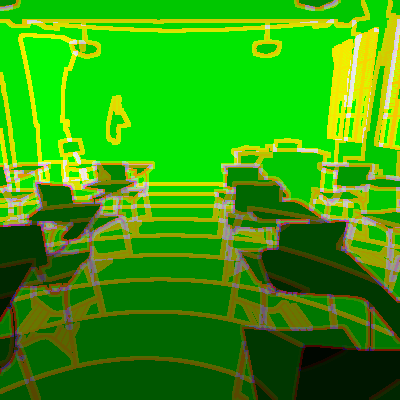}
		\caption{\classroom $K=5$}
	\end{subfigure}
	\begin{subfigure}[t]{.16\linewidth}
		\includegraphics[width=\linewidth]{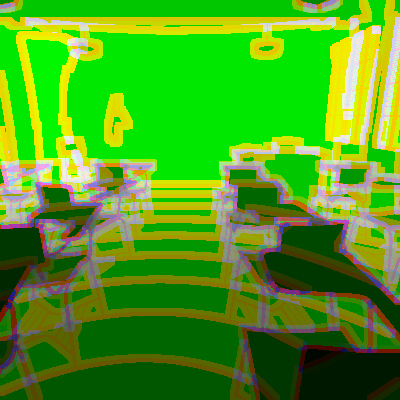}
		\caption{\classroom $K=9$}
	\end{subfigure}
	\caption{Visualization of the classified depth target for 16 classes along each ray and the largest classes mapped to $G$, $R$, $B$, \ie, a green value indicates a single class with the brightness encoding depth; a gray value means that all classes are at similar depth; and different colors in proximity indicate that classes vary. Small features are smoothed to neighboring pixels with increasing filter sizes (1-5-9).}
	\label{fig:classified_depth_examples}
\end{figure*}

To this end, we propose that the oracle is trained via classification, where each class corresponds to a discretized segment along the ray.
For every discrete ray segment, a high value indicates that it should receive (multiple) samples; a low value indicates that the segment can be skipped.
The surfaces are represented accurately, \ie,
\begin{equation}
	C_{x, y}(z) = 
	\begin{cases}
		1, & \text{if } d_z \leq d_s < d_{z+1} \\
		0, & \text{otherwise},
	\end{cases}
\end{equation}
where $C$ is the classification value, $d_s$ corresponds to the depth value of the first surface along the ray and $d_z$ and $d_{z+1}$ describe the discretization boundaries for the ray segment $z$.
This leads to a one-hot encoding as a target that can be trained using the common binary cross-entropy (BCE) loss. %

To further aid the depth oracle in predicting consistent outputs at depth discontinuities, we provide a multi-class target that is filtered in image-space and along depth. 
We blur depth values from neighboring pixels in the ground truth target.
To generate this filtered target, we use a radial (Euclidean distance) filter to include values of neighboring rays with a lower contribution:
\begin{equation}
	\hat C_{x, y}(z) =  \max_{i,j \, \in \pm \lfloor K/2 \rfloor} \left ( C_{x+i, y+j}(z) - \frac{\sqrt{i^2+j^2}}{\sqrt{2}\cdot \lfloor K/2 \rfloor} \right )  \label{eqn:pixel_neighborhood_filter} %
\end{equation}
where %
$K$ is the filter size. %
For example, using a $5\times5$ filter, rays at a distance of $3$, $2$, $1$, $0$ pixels contribute $0$, $\approx 0.30$, $\approx 0.65$, $1$ to the output, respectively.
For multiple depth values with the same discretization result, we only take the maximum result (contribution from the closest neighboring ray) to ensure a valid classification target as well as that the largest value is placed at the actual surface.

Following the idea of label smoothing, we also filter along the depth axis, using a simple 1D triangle filter with kernel size $Z$:
\begin{align}
	\grave C(z) &=  \text{min}\left( \sum_{i = -\lfloor Z/2 \rfloor}^{\lfloor Z/2 \rfloor} \hat  C(z + i) \frac{\lfloor Z/2 \rfloor + 1 - |i|}{\lfloor Z/2 \rfloor + 1}, 1 \right) 
	\label{eqn:depth_smoothing_filter}.
\end{align}

From a classification point of view, filtering decreases false negatives (at the cost of false positives) and thus
ensures that important regions are not missed.
This becomes apparent when considering that rays exist in a continuous space, \ie, the oracle does not need to learn hard boundaries  at depth discontinuities.
Translating a higher false positive rate to raymarching, we increase the likelihood for sampling regions that do not need any samples.
Thus, overall, we need to place more samples to hit the \emph{right} sample locations, while at the same time reducing the chance to miss surfaces.

Given that the oracle network serves a similar purpose as the coarse network in \gls{nerf} with the same capacity, while being evaluated only once instead of $64$ times, allowing the network to move towards false positive classifications is essential.
Additionally, false positives are handled by the local raymarching network, as empty space will still result in no image contribution, even at low sample counts.
A missed surface on the other hand would clearly reduce image quality.
Figure \ref{fig:classified_depth_filtering_drawing} shows an example slice for a fully filtered target and Figure \ref{fig:classified_depth_examples} shows visualizations for different filter sizes.
\revised{Note that the filtering only applies to the training targets for the depth oracle---no filtering is done during inference.}

\begin{figure}
	\centering
	\captionsetup{labelfont=bf,textfont=it}
	\includegraphics[width=\linewidth]{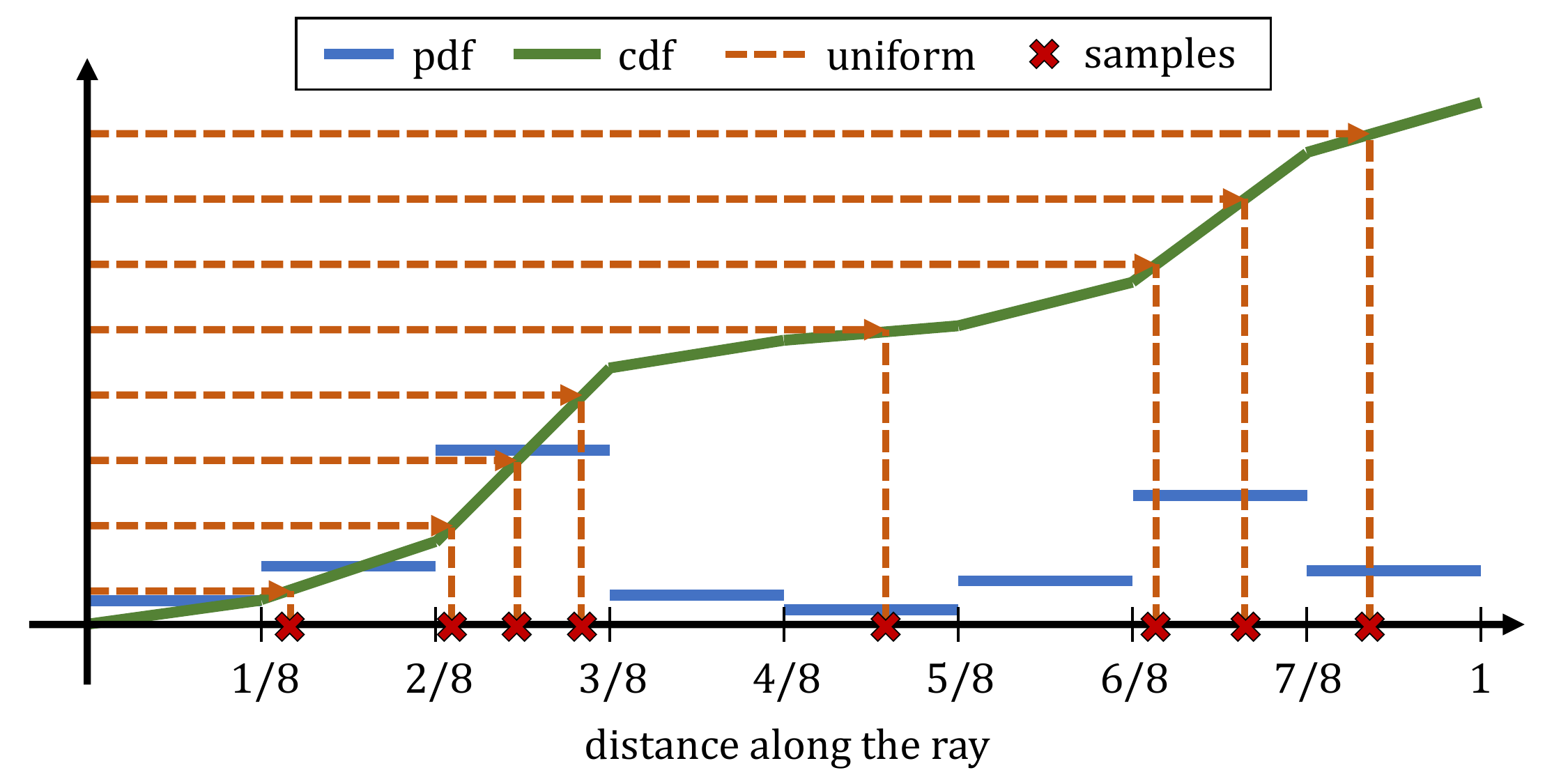}
	\caption{To place samples along the ray, we treat the classified depth output from the oracle as a piecewise-constant PDF (blue) and translate it into a CDF (green). Sampling the CDF at uniform steps (orange) concentrates samples (red $\times$) around regions with a high classification value.}
	\label{fig:sample_placement}
\end{figure}

Finally, to translate the classification outputs to sample locations, we use the same approach as \gls{nerf}~\cite{mildenhall2020nerf} when going from the coarse to the fine network.
\revised{Compared to \gls{nerf}, which builds a piecewise-constant probability density function (PDF) from the opacity outputs of the coarse shading network evaluated at multiple sample locations, we interpret our depth oracle output vector (which comes from a single oracle network evaluation) directly as a piecewise-constant PDF and similarly sample along the inverse transform of the corresponding cumulative distribution function (CDF); see Figure~\ref{fig:sample_placement}.}

\subsection{Ray Unification and Oracle Input}
\label{sec:SphPosDir}

\begin{figure}
	\centering
	\captionsetup{labelfont=bf,textfont=it}
	\begin{subfigure}[t]{.46\linewidth}
		\includegraphics[page=1,width=\linewidth]{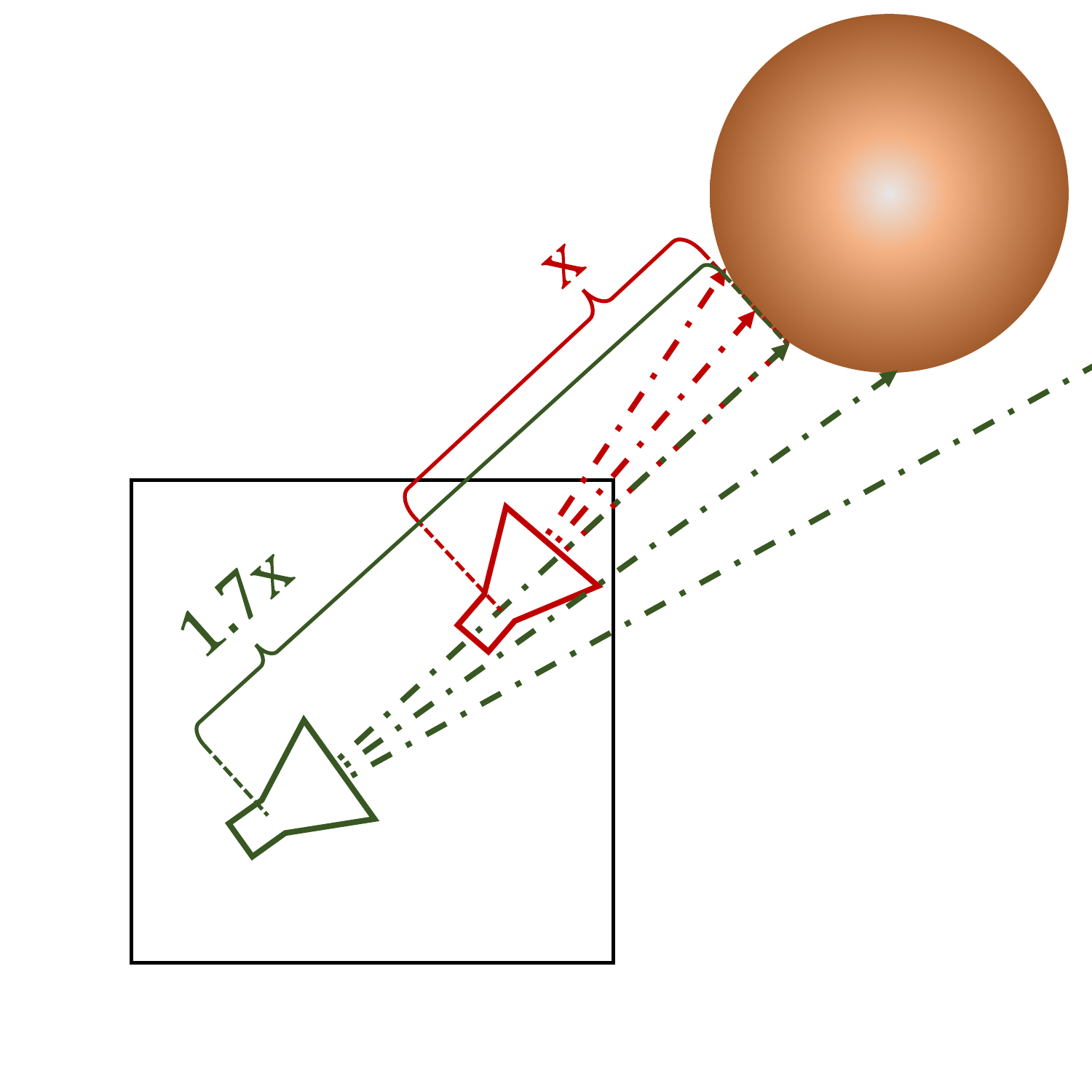}
	\end{subfigure}%
	\hfill%
	\begin{subfigure}[t]{.46\linewidth}
		\includegraphics[page=2,width=\linewidth]{figures/sphereposdir}
	\end{subfigure}
	
	\caption{Ray unification maps ray starting points to a sphere surrounding the view cell. Without ray unification (left) the same ray is encoded differently and requires different depth values; after ray unification (right) identical rays have identical depth values.}
	\label{fig:ray_unification}
\end{figure}

To take the burden of disambiguating rays originating from different starting locations from the oracle, we map every ray starting location onto a sphere circumscribed around the view cell, see Figure \ref{fig:ray_unification}.
\revised{This unification works well for arbitrary views looking outside the view cell. For 360$^\circ$ object captures, such as in the original \gls{nerf} work, our ray unification scheme be applied in an inverse manner, where the circumscribed sphere is placed around the object instead.}

To ease the task of the oracle network further, we supply 3D positions along the ray as input.
We place those at the centers of the discretized depth ranges and provide them as additional inputs to the first network layer.
We do not use positional or Fourier encoding for the depth oracle inputs, as this did not improve results in our experiments.
To work in unison with the shading network, for which we place samples logarithmically (Section~\ref{sec:nonlinear_sampling}), we also perform the same logarithmic transformation on the classification steps.

\subsection{Optimal Filter Sizes and Depth Oracle Inputs}
\label{sec:optimal_filter_size}
\revised{We conduct an ablation study to evaluate the impact of the filter sizes for our neighborhood filter $K$ (Equ.~\ref{eqn:pixel_neighborhood_filter}), depth smoothing filter $Z$ (Equ.~\ref{eqn:depth_smoothing_filter}) and the number of 3D input positions ($I=[1, 128]$).} %
We use the classified depth oracle network as the input for a locally raymarched shading network, similar to the experiment in Section~\ref{sec:localsampling}.

We vary the number of samples per ray $N = [2, 4, 8, 16]$ . %
Both the depth oracle network and the shading network contain $8$ hidden layers with $256$ hidden units. 
We first train the depth oracle network for \num{300000} iterations, before training the shading network for \num{300000} iterations using the oracle's predictions.
To illustrate the importance of our depth classification, we compare against a depth oracle that only predicts a single depth value with (\emph{SD unified}) and without (\emph{SD}) ray unification.
Our metric is the resulting quality of the RGB output of the shading network, which should improve when given better sample positions by the depth oracle.

The results in Table~\ref{tbl:filter_ablation} show that (1) ray unification adds about \revised{\SI{0.6}{\decibel}} in PSNR, (2) using a classification network adds another \revised{\SIrange{0.6}{1.5}{\decibel}}, (3) providing multiple samples along the ray as input adds \revised{\SI{1}{\decibel}}, (4) the neighborhood filter adds \revised{\SIrange{1.3}{1.8}{\decibel}}, and the depth smoothing filter adds \revised{\SI{0.1}{\decibel}}.
These improvements come at no inference cost (filtering) or virtually no inference cost (ray unification, multi input). %
In total, our additions improve the PSNR by \revised{\SIrange{3.3}{4.7}{\decibel}}.
Note that an even larger filter size may reduce overall quality, as many samples are placed in empty space rather than on the surface.

\def\rtbs{\hspace{2.0pt}}
\def\rtbsx{\hspace{4.0pt}}
\def\rtbm{\hspace{4.5pt}}
\begin{table}[t]\centering
	\captionsetup{labelfont=bf,textfont=it}
	\caption{PSNR and FLIP results averaged over all scenes, evaluating the neighborhood filter size (K-$X$), the depth smoothing filter size (Z-$X$) and the number of inputs for the depth oracle (I-$X$) over the number of samples per ray $N$ used for raymarching. We also compare against an oracle that only predicts a single depth value (\emph{SD}), as well as a single depth value with unified input (\emph{SD unified}). %
	}\label{tbl:filter_ablation}
	\resizebox{\linewidth}{!}{
	\begin{tabular}{@{\rtbs}l@{\rtbm}@{\rtbm}c@{\rtbs}@{\rtbs}c@{\rtbs}@{\rtbs}c@{\rtbs}@{\rtbs}c@{\rtbm}@{\rtbm}c@{\rtbs}@{\rtbs}c@{\rtbs}@{\rtbs}c@{\rtbs}@{\rtbs}c@{\rtbs}@{\rtbs}c@{\rtbs}}\toprule
		&\multicolumn{4}{c}{PSNR $\uparrow$} &\multicolumn{4}{c}{FLIP $\downarrow$} \\\cmidrule{1-9}
Method\hspace{6pt}\textbackslash\hspace{6pt}  N &2 &4 &8 &16 &2 &4 &8 &16 \\\midrule
SD &26.686 &27.401 &28.220 &29.085 &0.092 &0.084 &0.078 &0.073 \\
SD unified &27.423 &28.052 &28.825 &29.554 &0.085 &0.079 &0.074 &0.071 \\
K-1 Z-1 I-1 &27.325 &28.697 &30.068 &31.145 &0.082 &0.073 &0.065 &0.061 \\
K-5 Z-1 I-1 &28.685 &30.521 &31.988 &32.982 &0.075 &0.066 &0.059 &0.055 \\
K-5 Z-1 I-128 &29.956 &31.746 &32.951 &33.760 &0.067 &\textbf{0.061} &\textbf{0.056} &\textbf{0.053} \\
K-5 Z-5 I-128 &\textbf{30.071} &\textbf{31.842} &33.027 &33.836 &\textbf{0.067} &\textbf{0.061} &\textbf{0.056} &\textbf{0.053} \\
K-9 Z-1 I-1 &28.831 &30.881 &32.617 &33.495 &0.075 &0.065 &0.057 &0.055 \\
K-9 Z-1 I-128 &29.299 &31.645 &\textbf{33.125} &\textbf{33.994} &0.071 &\textbf{0.061} &\textbf{0.056} &\textbf{0.053} \\
K-9 Z-9 I-128 &28.737 &30.847 &32.261 &33.302 &0.076 &0.066 &0.060 &0.056 \\
\bottomrule
	\end{tabular}
}
\end{table}

\section{Evaluation}
\label{sec:evaluation}
To evaluate \ours, we focus on three competing requirements of (neural) scene representations: \emph{quality} of the generated images, \emph{efficiency} of the image generation, and \emph{compactness} of the representation.
Clearly tradeoffs between them are possible, but an ideal representation should generate high-quality outputs in real-time, while being compact and extensible, \eg, for streaming dynamic scenes.%
We compare against \gls{nerf}~\cite{mildenhall2020nerf}, \gls{nsvf}~\cite{liu2020neural}, \gls{llff}~\cite{mildenhall2019llff} and \gls{nex}~\cite{Wizadwongsa2021NeX} to evaluate methods that choose different tradeoffs among our three goals. %

These methods capture a mix between being strictly \gls{mlp}-based (\gls{nerf}), using explicit structures and \glspl{mlp} (\gls{nsvf}, \gls{nex}) and using a mostly image-based representation (\gls{llff}).
For \gls{nex}, we include an additional variant that does not bake radiance coefficients and neural basis functions into an \gls{mpi}, but recomputes those via \gls{mlp} inference during test time (\gls{nex}-\gls{mlp}).
For \gls{nsvf}, we run a grid search and evaluate on three representative variants (\emph{\gls{nsvf}-small}, \emph{\gls{nsvf}-medium} and \emph{\gls{nsvf}-large}) that capture the lowest memory footprint, best quality-speed tradeoff, and best quality respectively.
Furthermore, we include a variant of \gls{nerf} that uses our \emph{log+warp} sampling to show the effect of the sampling strategy in isolation (\gls{nerf} (log+warp)).
See Appendix \ref{app:baselines} for details about the methods.

We analyze the ability to extract novel high-quality views for generated content where reference depth maps are available during training.
As an additional proof-of-concept, we extract estimated depth maps from a densely sampled \gls{nerf} for each scene, and use these depth maps to train our depth oracles, showcasing a solution for scenes without available ground truth depth.
We evaluate quality by computing PSNR and FLIP~\cite{andersson_2020_flip} against ground truth renderings, efficiency as FLOP per pixel and compactness by total storage cost for the representation.
For all methods, images are downsampled to a resolution of $400\times 400$ to speed up training.

\def\rtbst{\hspace{7.0pt}}

\newcommand{\reducedstrut}{\vrule width 0pt height 1.0\ht\strutbox depth 1.0\dp\strutbox\relax}
\newcommand{\topa}[1]{%
	\begingroup
	\setlength{\fboxsep}{1pt}%
	\colorbox[HTML]{29227d}{\reducedstrut\textcolor[HTML]{FFFFFF}{\textbf{#1}}\/}%
	\endgroup
}
\newcommand{\topb}[1]{%
	\begingroup
	\setlength{\fboxsep}{1pt}%
	\colorbox[HTML]{5950cf}{\reducedstrut\textcolor[HTML]{FFFFFF}{{#1}}\/}%
	\endgroup
}
\newcommand{\topc}[1]{%
	\begingroup
	\setlength{\fboxsep}{1pt}%
	\colorbox[HTML]{d0cdf1}{\reducedstrut\textcolor[HTML]{000000}{{#1}}\/}%
	\endgroup
}

\begin{table*}[t]\centering
	\captionsetup{labelfont=bf,textfont=it}
	\caption{Across all our test scenes, \ours shows a significant improvement in quality-speed tradeoff, beating \gls{nerf} in most scenes with only $2-4$ samples per ray, resulting in $48\times$ to $78\times$ fewer FLOP per pixel at equal or better quality. Where \gls{nsvf} struggles to achieve a consistent quality or performance for our tested scenes, especially for larger scenes such as \forest, \ours performs consistently well across all scales. \gls{llff} is the cheapest to compute, but fails to replicate the quality of the other methods, in addition to requiring storage that scales unfavorably with the amount of training images. Although \gls{nex} is also very efficient at rendering, it suffers from artifacts related to the \gls{mpi} representation when poses differ too much from the reference pose, and requires roughly $24\times$ the amount of storage to achieve worse quality than \ours across the board. \gls{nex}-\gls{mlp} is able to remedy some of the artifacts at a cost of $9\times$ worse performance compared to \ours-4. Finally, using depth maps extracted from a dense \gls{nerf} without depending on available ground truth depth, \ours-noGT still achieves the best tradeoff between performance, quality and storage overall. \revised{Top results in each column are color coded as \topa{Top 1}, \topb{Top 2} and \topc{Top 3}.}}\label{tbl:main_results}
	\begin{tabular}{@{\rtbs}l@{}@{}r@{\rtbs}@{\rtbs}r@{\rtbst}@{\rtbs}r@{}@{\rtbs}r@{\rtbst}@{\rtbs}r@{}@{\rtbs}r@{\rtbst}@{\rtbs}r@{}@{\rtbs}r@{\rtbst}@{\rtbs}r@{}@{\rtbs}r@{\rtbst}@{\rtbs}r@{}@{\rtbs}r@{\rtbst}@{\rtbs}r@{}@{\rtbs}r@{\rtbst}@{\rtbs}r@{}@{\rtbs}r@{\rtbs}}\toprule
		& & &\multicolumn{2}{c}{\sanmiguel} &\multicolumn{2}{c}{\pavillon} &\multicolumn{2}{c}{\classroom} &\multicolumn{2}{c}{\bulldozer} &\multicolumn{2}{c}{\forest} &\multicolumn{2}{c}{\barbershop}  &\multicolumn{2}{c}{Average} \\\midrule
		Method &\shortstack{Storage\\ {[MiB]}} &\shortstack{MFLOP\\ per pixel} &PSNR &FLIP &PSNR &FLIP &PSNR &FLIP &PSNR &FLIP &PSNR &FLIP &PSNR &FLIP &PSNR &FLIP \\\midrule
		\ours-2                 &\topc{3.6} &\topc{2.70}  &26.01        &.094         &30.50        &.103         &31.66        &.061         &30.15        &.063         &29.29        &.082         &29.41        &.074         &29.50        &.079         \\
		\ours-2-noGT            &\topc{3.6} &\topc{2.70}  &25.33        &.098         &29.84        &.103         &30.11        &.067         &26.92        &.077         &28.36        &.089         &29.01        &.075         &28.26        &.085         \\
		\ours-4                 &\topc{3.6} &4.36         &27.41        &.080         &31.07        &.098         &33.43        &.058         &33.46        &.048         &30.63        &\topc{.077}  &30.84        &.065         &31.14        &.071         \\
		\ours-4-noGT            &\topc{3.6} &4.36         &26.19        &.090         &30.69        &.096         &31.44        &.061         &29.78        &.060         &29.31        &.086         &30.42        &.067         &29.64        &.077         \\
		\ours-8                 &\topc{3.6} &7.66         &28.65        &\topc{.071}  &31.46        &.096         &\topb{35.23} &\topc{.048}  &35.88        &\topc{.039}  &\topa{32.09} &\topa{.070}  &31.72        &.060         &\topc{32.50} &\topb{.064}  \\
		\ours-8-noGT            &\topc{3.6} &7.66         &26.88        &.086         &31.56        &\topc{.091}  &33.19        &.055         &32.96        &.047         &29.98        &.084         &31.73        &.062         &31.05        &.071         \\
		\ours-16                &\topc{3.6} &14.29        &\topb{29.67} &\topb{.065}  &\topc{31.79} &.094         &\topa{36.27} &\topa{.045}  &\topa{36.98} &\topa{.036}  &\topb{31.32} &\topb{.074}  &32.15        &.059         &\topa{33.03} &\topa{.062}  \\
		\ours-16-noGT           &\topc{3.6} &14.29        &27.70        &.078         &\topb{32.22} &\topa{.088}  &34.63        &.049         &35.41        &.040         &\topc{30.74} &.079         &\topc{32.80} &\topc{.057}  &32.25        &\topc{.065}  \\\midrule
		\gls{nerf}              &\topb{3.2} &211.42       &25.19        &.117         &29.54        &.115         &34.02        &.056         &\topb{36.83} &\topb{.038}  &23.90        &.151         &\topa{33.63} &\topa{.052}  &30.52        &.088         \\
		\gls{nerf} (log + warp) &\topb{3.2} &211.42       &\topc{28.98} &.074         &\topa{32.88} &\topb{.089}  &\topc{35.19} &.051         &\topc{36.22} &.040         &28.97        &.101         &\topb{33.60} &\topb{.055}  &\topb{32.64} &.068         \\\midrule
		\gls{nsvf}-small        &\topa{2.3} &74.66        &24.00        &.132         &29.42        &.110         &31.00        &.070         &25.75        &.167         &23.79        &.159         &27.72        &.094         &26.95        &.122         \\
		\gls{nsvf}-medium       &4.6        &132.03       &25.07        &.110         &29.81        &.105         &33.04        &.055         &26.51        &.163         &25.08        &.135         &29.62        &.077         &28.19        &.108         \\
		\gls{nsvf}-large        &8.3        &187.52       &25.73        &.097         &30.48        &.099         &34.06        &.051         &33.14        &.042         &26.05        &.119         &30.61        &.061         &30.01        &.078         \\\midrule
		\gls{llff}              &4130.6     &\topa{.03}   &24.53        &.106         &27.50        &.123         &24.87        &.114         &24.76        &.114         &22.19        &.148         &24.13        &.129         &24.66        &.122         \\\midrule
		\gls{nex}               &88.8       &\topb{.06}   &28.07        &.094         &26.28        &.174         &30.34        &.085         &29.20        &.072         &20.95        &.220         &22.98        &.152         &26.30        &.133         \\
		\gls{nex}-\gls{mlp}     &89.0       &42.71        &\topa{30.68} &\topa{.060}  &30.41        &.102         &34.10        &\topb{.046}  &34.03        &.046         &24.65        &.125         &29.45        &.075         &30.55        &.076         \\
		\bottomrule
	\end{tabular}
\end{table*}

\newcommand\imgsize{0.15\linewidth}
\begin{figure*}
	\centering
	\captionsetup{labelfont=bf,textfont=it}
	\begin{subfigure}{\linewidth}
		\centering
		\begin{minipage}{\linewidth}
			\centering
			\hspace{0.74pt}
			\includegraphics[width=\imgsize]{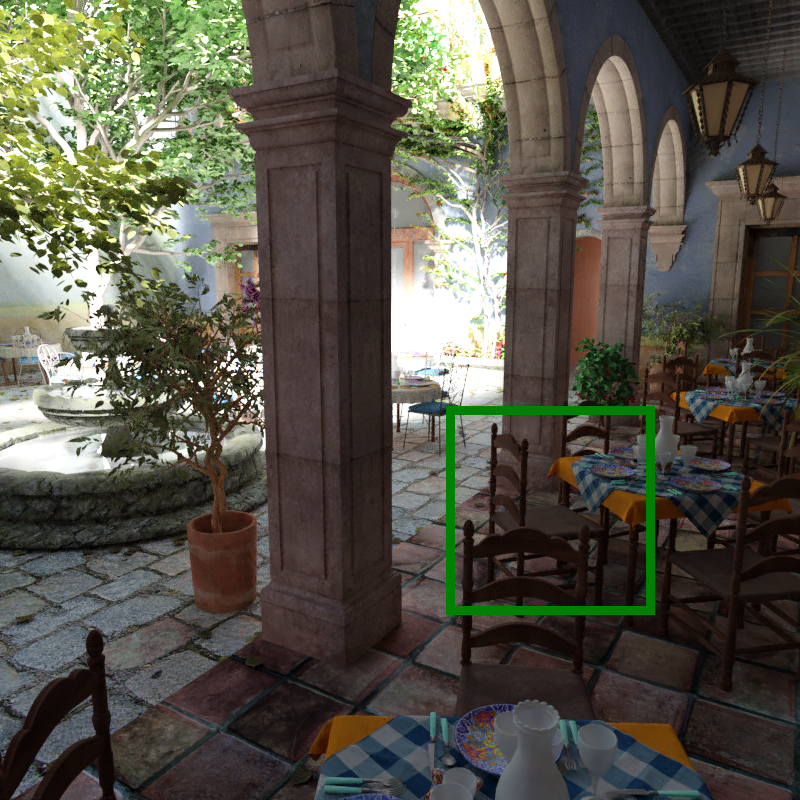}
			\includegraphics[width=\imgsize]{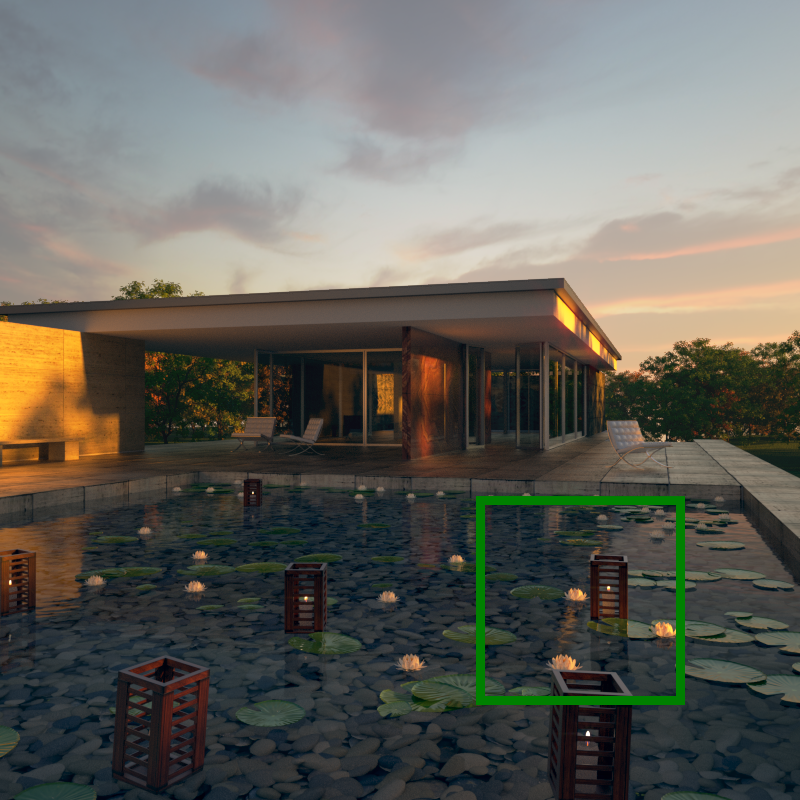}
			\includegraphics[width=\imgsize]{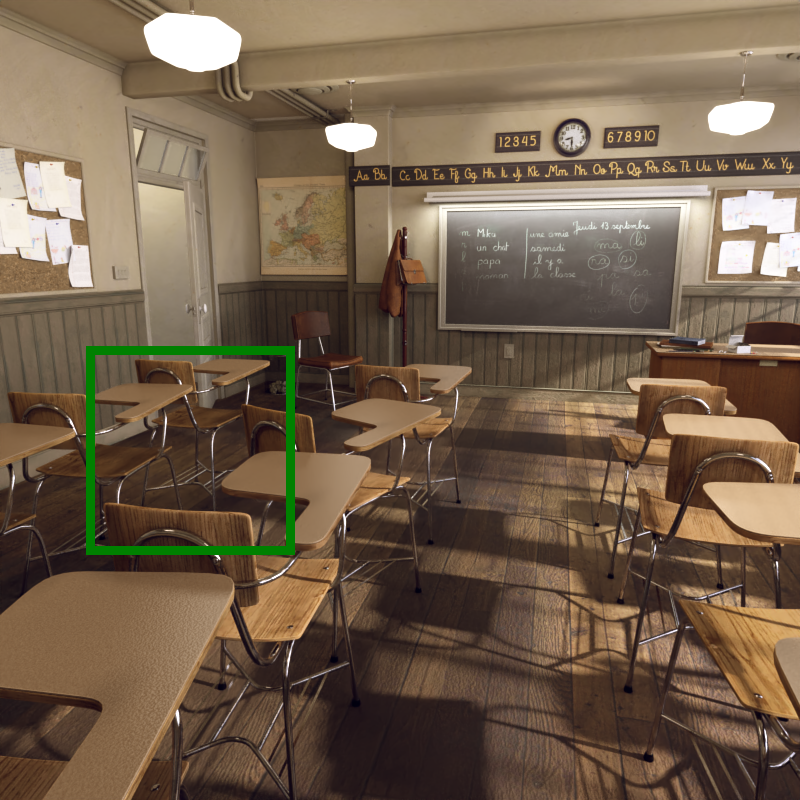}
			\includegraphics[width=\imgsize]{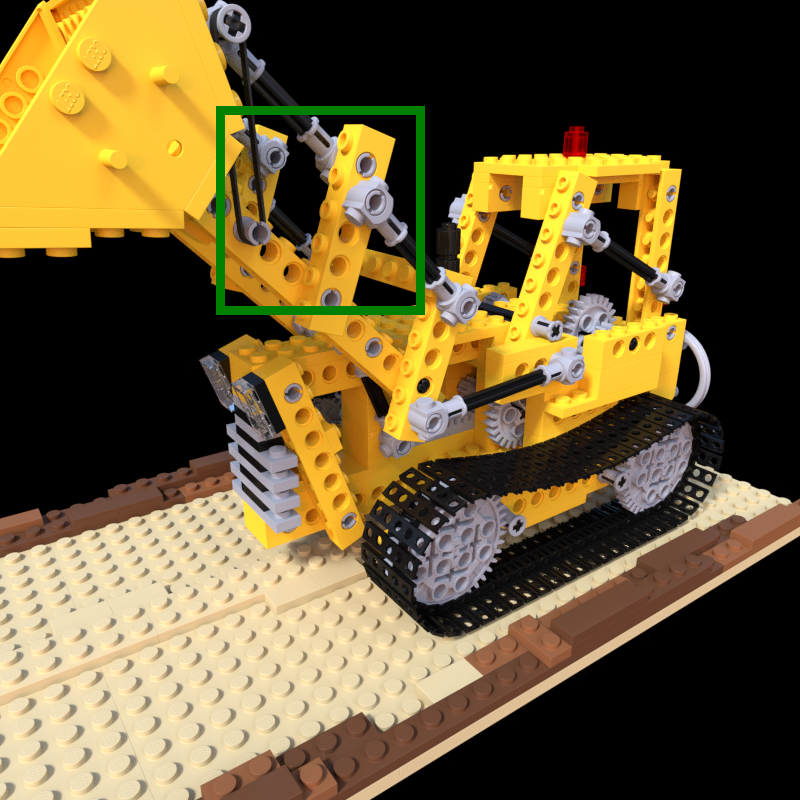}
			\includegraphics[width=\imgsize]{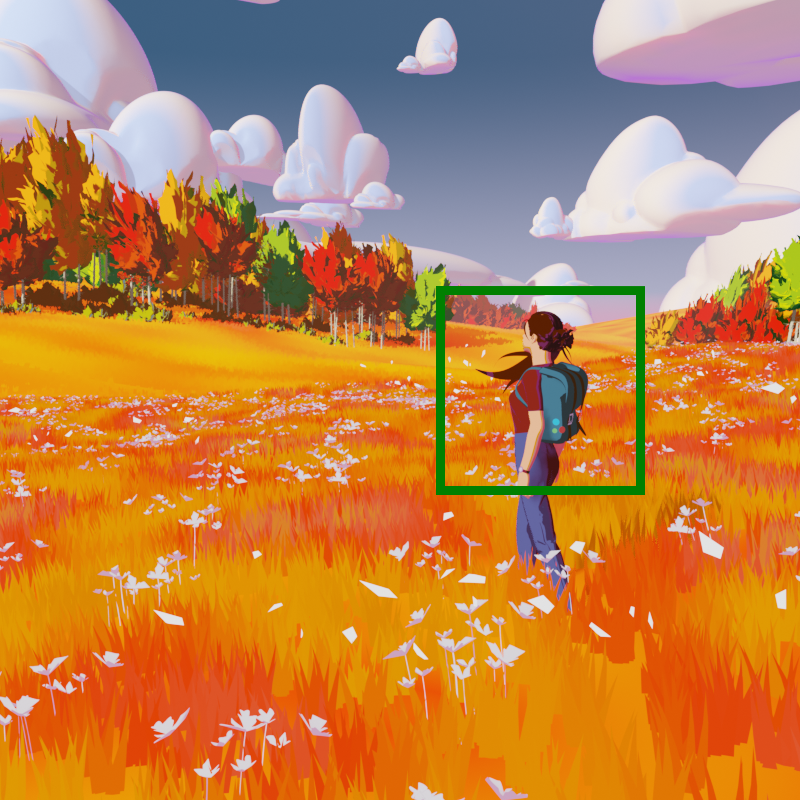}
			\includegraphics[width=\imgsize]{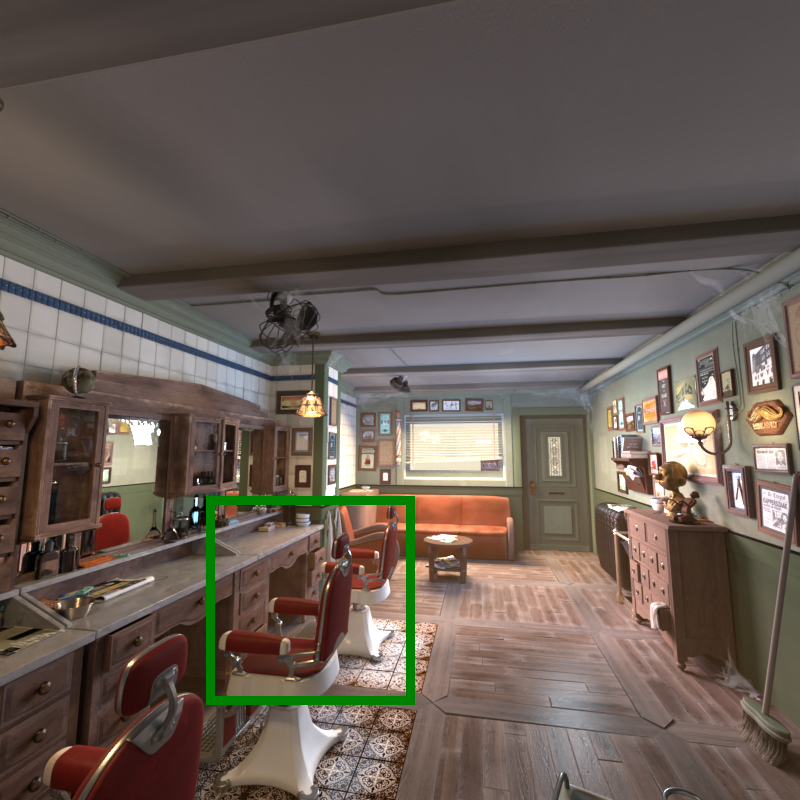}
			\hfill
		\end{minipage}
		\caption*{Ground Truth}
		\hfill
	\end{subfigure}
	\hfill
	\begin{subfigure}[t]{\imgsize}
		\begin{minipage}{\linewidth}
			\includegraphics[width=\linewidth]{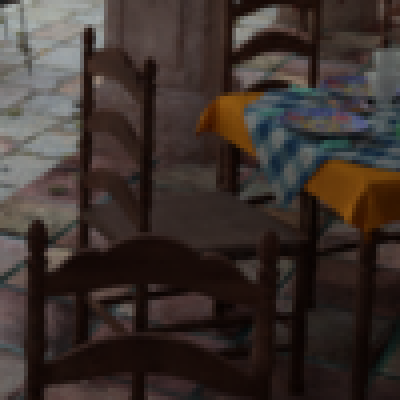}
			\includegraphics[width=\linewidth]{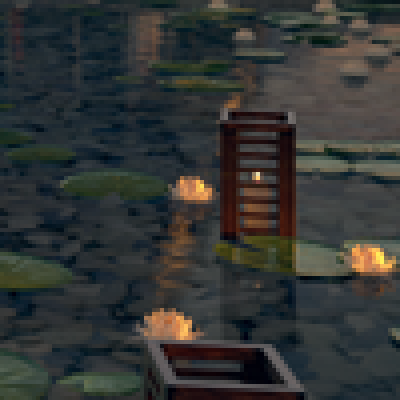}
			\includegraphics[width=\linewidth]{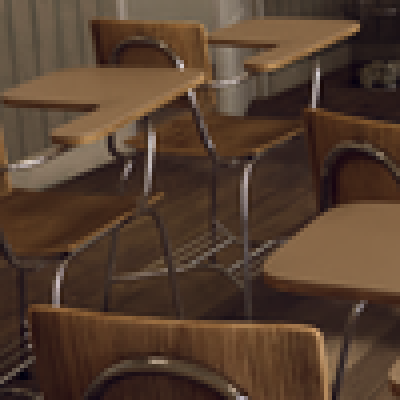}
			\includegraphics[width=\linewidth]{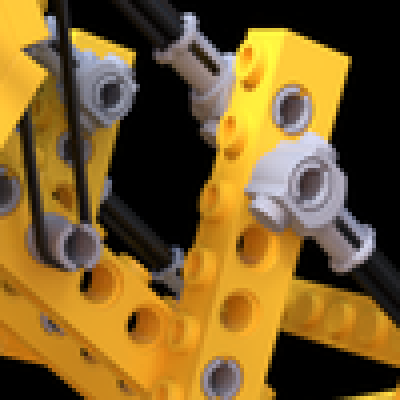}
			\includegraphics[width=\linewidth]{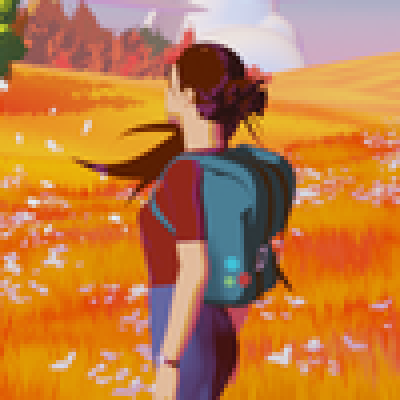}
			\includegraphics[width=\linewidth]{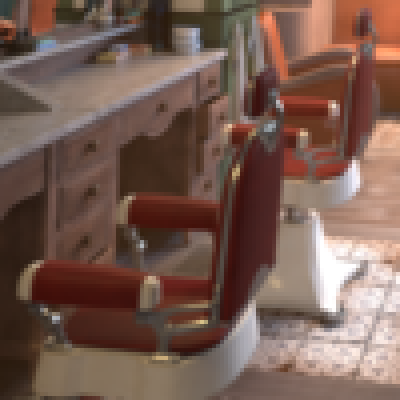}
		\end{minipage}
		\caption{Ground Truth}
	\end{subfigure}
	\begin{subfigure}[t]{\imgsize}
		\begin{minipage}{\linewidth}
			\includegraphics[width=\linewidth]{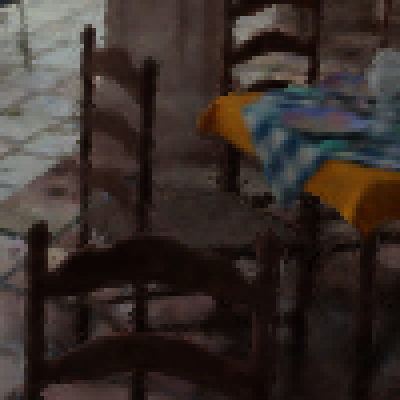}
			\includegraphics[width=\linewidth]{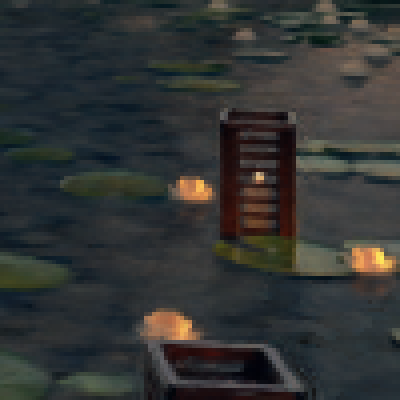}
			\includegraphics[width=\linewidth]{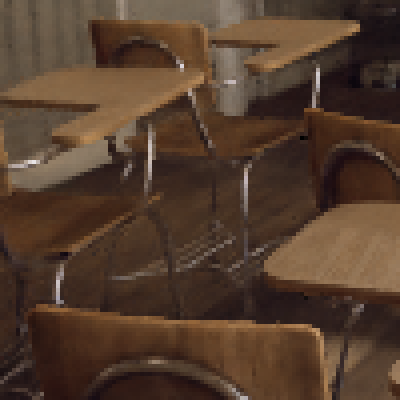}
			\includegraphics[width=\linewidth]{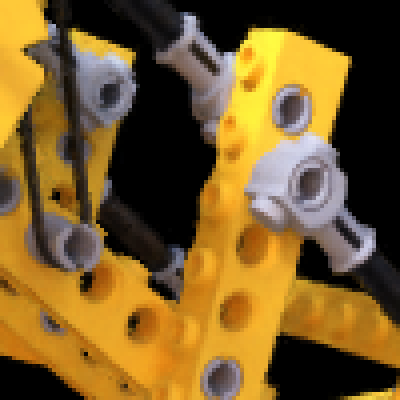}
			\includegraphics[width=\linewidth]{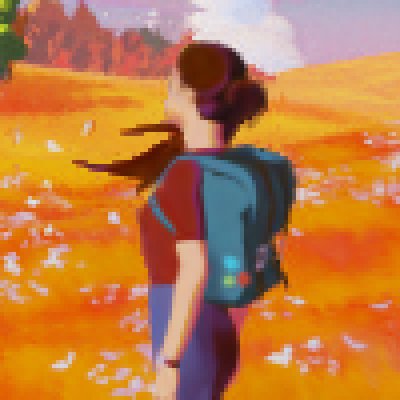}
			\includegraphics[width=\linewidth]{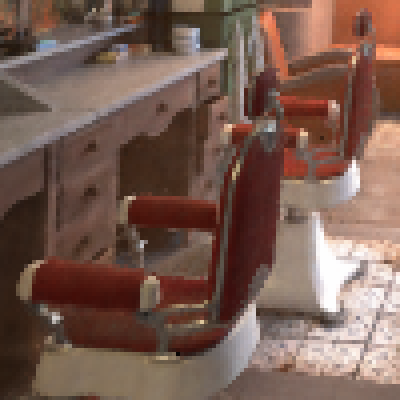}
		\end{minipage}
		\caption{\ours-4}
	\end{subfigure}
	\begin{subfigure}[t]{\imgsize}
		\begin{minipage}{\linewidth}
			\includegraphics[width=\linewidth]{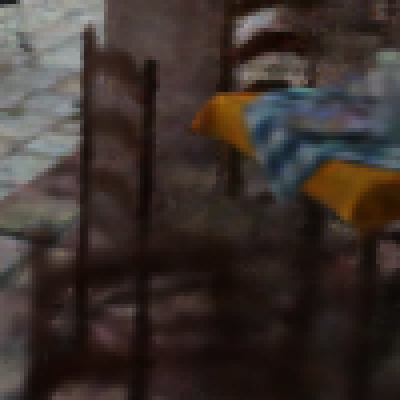}
			\includegraphics[width=\linewidth]{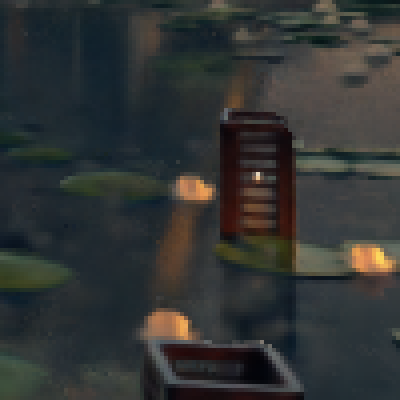}
			\includegraphics[width=\linewidth]{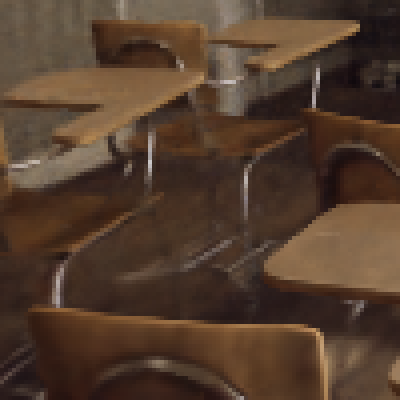}
			\includegraphics[width=\linewidth]{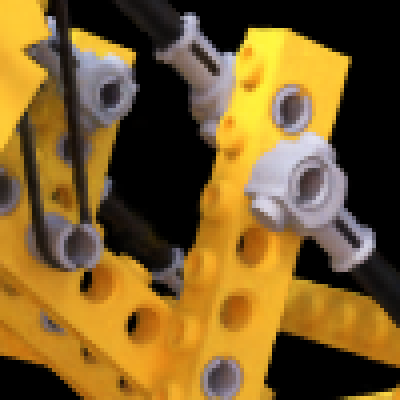}
			\includegraphics[width=\linewidth]{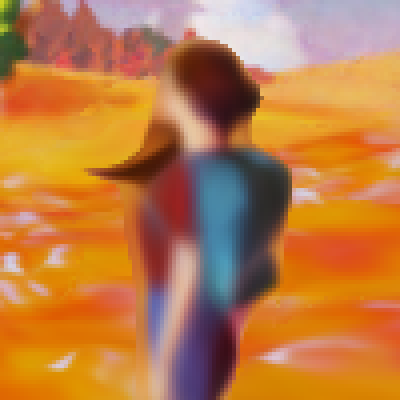}
			\includegraphics[width=\linewidth]{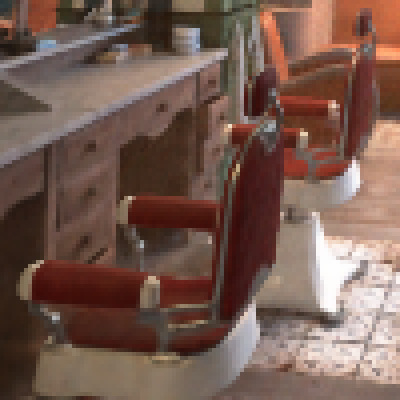}
		\end{minipage}
		\caption{\gls{nerf}}
	\end{subfigure}
	\begin{subfigure}[t]{\imgsize}
		\begin{minipage}{\linewidth}
			\includegraphics[width=\linewidth]{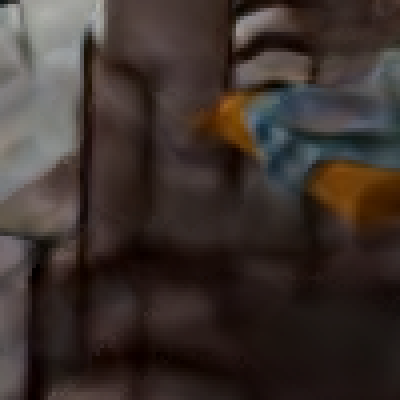}
			\includegraphics[width=\linewidth]{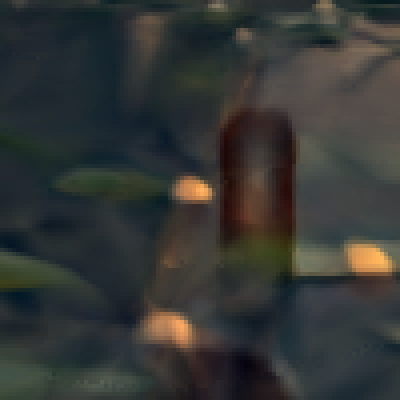}
			\includegraphics[width=\linewidth]{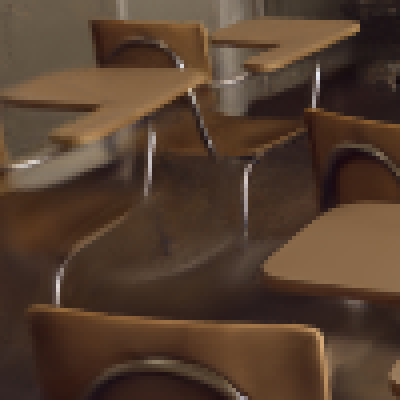}
			\includegraphics[width=\linewidth]{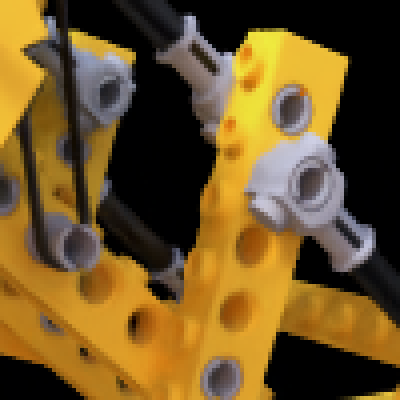}
			\includegraphics[width=\linewidth]{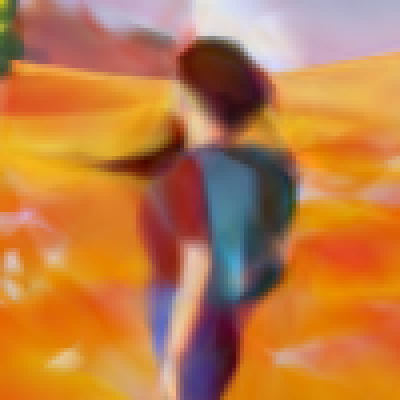}
			\includegraphics[width=\linewidth]{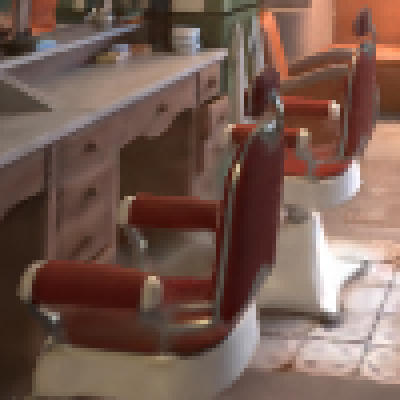}
		\end{minipage}
		\caption{\gls{nsvf}-medium}
	\end{subfigure}
	\begin{subfigure}[t]{\imgsize}
		\begin{minipage}{\linewidth}
			\includegraphics[width=\linewidth]{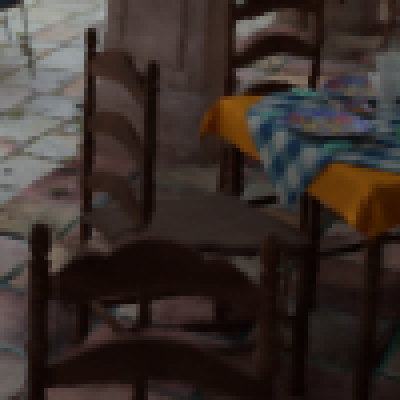}
			\includegraphics[width=\linewidth]{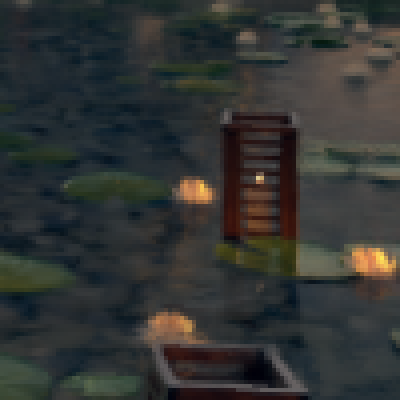}
			\includegraphics[width=\linewidth]{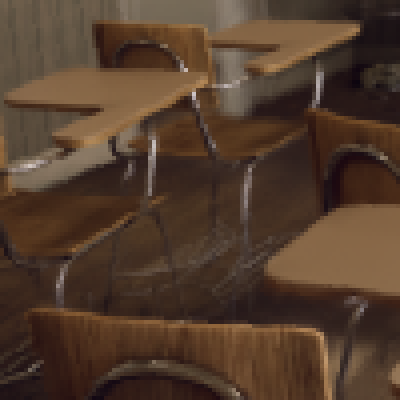}
			\includegraphics[width=\linewidth]{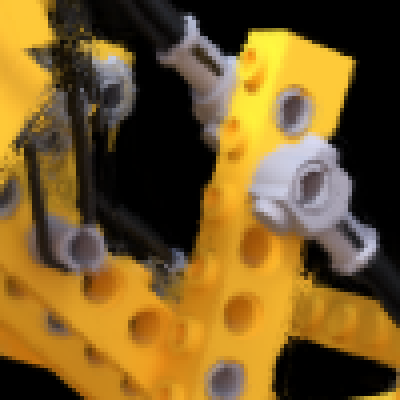}
			\includegraphics[width=\linewidth]{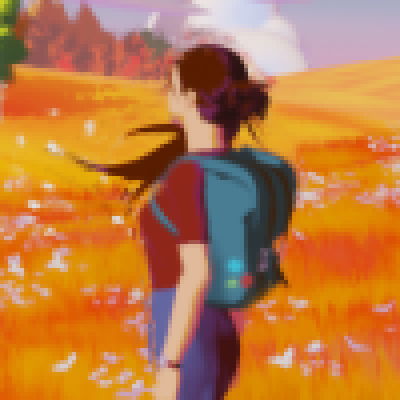}
			\includegraphics[width=\linewidth]{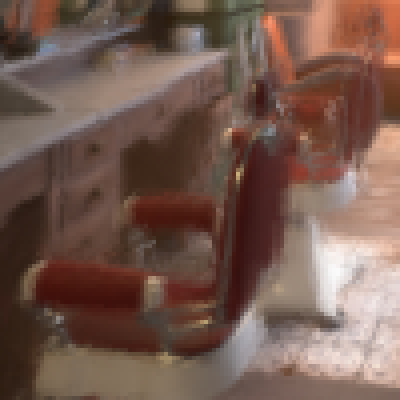}
		\end{minipage}
		\caption{\gls{llff}}
	\end{subfigure}
	\begin{subfigure}[t]{\imgsize}
		\begin{minipage}{\linewidth}
			\includegraphics[width=\linewidth]{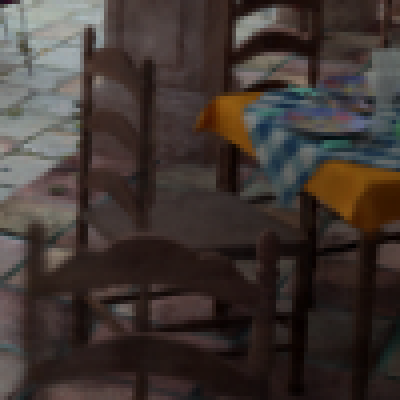}
			\includegraphics[width=\linewidth]{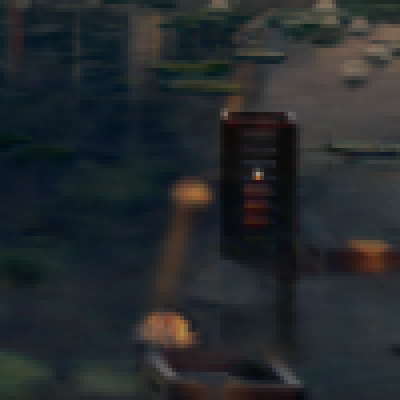}
			\includegraphics[width=\linewidth]{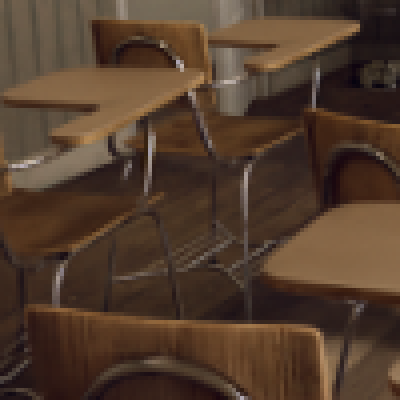}
			\includegraphics[width=\linewidth]{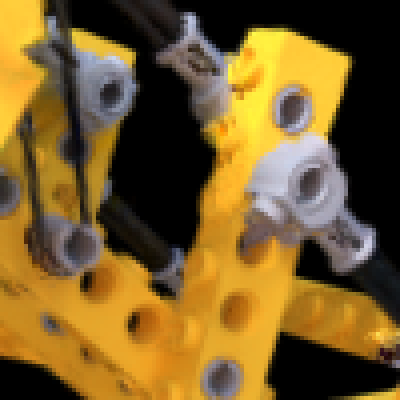}
			\includegraphics[width=\linewidth]{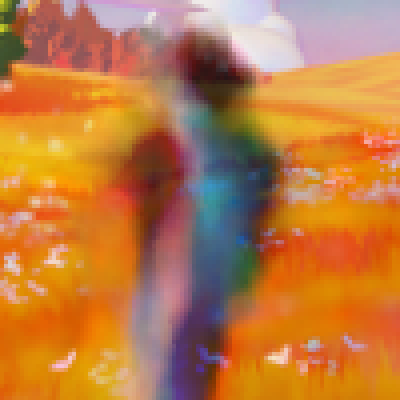}
			\includegraphics[width=\linewidth]{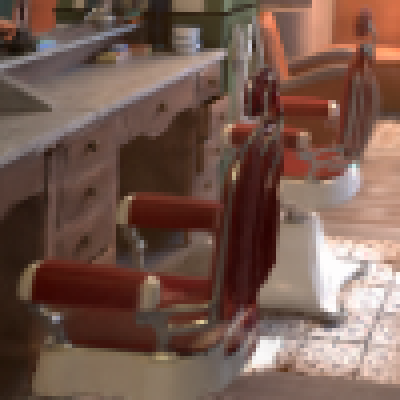}
		\end{minipage}
		\caption{\gls{nex}}
	\end{subfigure}
	\caption{Even on our challenging dataset, \ours achieves higher quality on average than all other methods at only 4 samples per ray.
	\gls{nerf} manages to faithfully reconstruct smaller scenes such as \bulldozer and \barbershop, but struggles with large-scale scenes such as \forest or highly complex geometry such as in \classroom. \gls{nsvf} shows worse quality on average compared to \ours-4 at increased memory requirements and much worse performance. Although \gls{llff} requires significantly more memory, it still struggles to represent fine details accurately, even at a cropped field of view. \gls{nex} achieves good quality for \sanmiguel, but larger rotations and offsets from its reference pose cause significant artifacts due to its explicit \gls{mpi} representation.}
	\label{fig:main_eval_qualitative}
\end{figure*}

\subsection{Training and Real-time Implementation}
We use K-$5$ Z-$5$ I-$128$ for our depth oracle network (see Section~\ref{sec:optimal_filter_size}) and use various numbers of samples per ray, named \ours-$X$, where $X$ denotes the number of samples per ray.
We transform each sample logarithmically and warp it towards the view cell, as described in Section~\ref{sec:nerf_sampling}.
We train each network for \num{300000} iterations and use the checkpoint with the lowest validation loss for testing. 
We use Adam~\cite{kingma2015adam} with a learning rate of $0.0005$ and a batch size of $4096$ samples per iteration.

For the RGB output of the shading network we use standard MSE loss, while the depth classification uses BCE loss.
\revised{Furthermore, during initial experiments, we observed that shading networks with low sample counts tend to ``cheat'' in their outputs, relying on the opacity outputs to mix the background color into the accumulation, resulting in a darkening with black background color. 
At low sample counts, this tends to hurt generalization, especially when rays \emph{should} accumulate to an opacity of at least $1$ for opaque surfaces.
As a result, dark pixel artifacts are visible in some test set views, which we remedy by adding an additional \emph{opacity loss} term that forces the accumulated ray opacity $\delta$ to be \emph{at least} $1$:
\begin{equation}
	\text{loss}_{O} =
	\begin{cases}
		0, & \text{if } \sum_{i=1}^{X} \delta_i \geq 1 \\
		\left( \left( \sum_{i=1}^{X} \delta_i \right) - 1 \right)^2, & \text{otherwise},
	\end{cases}
\end{equation}
where $X$ is again the number of samples per ray. 
Our final loss function is a weighted sum of both the MSE and the opacity loss
\begin{equation}
	\text{loss} = \alpha \cdot \text{loss}_{\text{MSE}} + \beta \cdot \text{loss}_{O}.
\end{equation}

For all \ours experiments, we use $\alpha = 1.0$ and $\beta = 10.0$ to conservatively remove all dark pixel artifacts.
Lower values for $\beta$ could be selectively applied to further increase quality in certain scenes. }
The network architecture of \ours is visualized in Appendix~\ref{app:ours}.
For \ours without ground truth depth (\ours-$X$-noGT), we train a dense \gls{nerf}  with $12$ layers of $512$ hidden units each using $128$ samples per ray, and extract depth maps for all poses. %
Using these depth maps, we train \ours in exactly the same way as when using the reference depth maps.

For our prototype real-time implementation of \ours, we use a combination of TensorRT and CUDA.
Ray unification, space warping, feature generation and raymarching run in custom CUDA kernels with one thread per ray.
Both networks are evaluated using TensorRT in half floating point precision.
All parts still offer significant optimization potential, by, \eg, using multiple threads during input feature generation or reducing the bit length for the network evaluation further. 

Nevertheless, we already achieve real-time rendering at medium resolutions on an NVIDIA RTX 3090.
Ray unification and first feature generation (\SI{0.21}{\ms}), the oracle network (\SI{12.64}{\ms}), space warping and the second feature generation (\SI{3.46}{\ms}), the shading network (\SI{34.9}{\ms}), and final color generation (\SI{0.09}{\ms}) take a total of \SI{51.3}{\ms} for a $800\times 800$ image and $2$ samples per ray, \ie, about $20$ frames per second.
Note that the shading network still is the major cost in our setup and that computation times increase nearly linearly with the number of samples: 
\SI{34.9}{\ms}, \SI{65.4}{\ms}, \SI{136.4}{\ms} and \SI{270.4}{\ms} for $2$, $4$, $8$ and $16$ samples, respectively.

\subsection{Dataset} 
\label{sec:datasets}
We collect a set of scenes that showcase both fine, high-frequency details as well as large depth ranges to validate that \ours is applicable to a wide variety of scenes. 
All datasets are rendered using Blender, using their Cycles path tracer to render $300$ images for each scene (taking approximately 20 minutes per image on a 64 core CPU), which we split into train/validation/test sets at a $70\% / 10\% / 20\%$ ratio.
For each scene, we define a view cell that showcases a wide range of movement and rotation to reveal disocclusions and challenging geometry, while not intersecting geometry within the scene.
Poses are randomly sampled within the view cell, limiting the rotation to \SI{30}{degrees} in pitch and \SI{20}{degrees} in yaw relative to the initial camera direction.
We limit the rotation to be comparable to plane-based representations such as \gls{nex} and \gls{llff}---\ours would be able to handle unrestricted rotation angles.
Please refer to Appendix \ref{app:scenes} for more details about the scenes.

\subsection{Comparisons}

The results of our main evaluation are summarized in Table~\ref{tbl:main_results} and qualitative example views are shown in Figure~\ref{fig:main_eval_qualitative}.
In the following we individually compare the competing approaches to \ours in terms of \emph{quality}, \emph{efficiency}, and \emph{compactness}.

\paragraph*{\gls{nerf}}
Compared to \gls{nerf}, the advantages of both our sampling strategy and oracle network are immediately visible.
\revised{On average, \ours achieves equal or better quality with only $2-4$ samples per ray compared to \gls{nerf}'s 256.}
At $16$ samples, \ours achieves up to \SI{7}{\decibel} higher PSNR, and outperforms \gls{nerf} on all scenes except for \barbershop.
At the same time, \ours is \numrange{15}{78}$\times$ faster to evaluate than \gls{nerf}.
As both only use two \glspl{mlp}, they are among the most compact methods---the small increase in memory of \ours is due to the increased number of inputs and outputs of our oracle network compared to the coarse \gls{nerf} network.
\revised{Even when applying our improved sampling strategy for \gls{nerf} (log+warp)},  \ours achieves equal quality with just $8$ samples.
Overall, \ours is superior compared to \gls{nerf} ---especially for large open scenes---both when considering \emph{quality}, \emph{efficiency} and \emph{compactness} as a whole, and when considering them individually.

\paragraph*{\gls{nsvf}}
The comparison to \gls{nsvf} is interesting, as \gls{nsvf} is able to adjust the tradeoff between quality, efficiency and compactness by changing the voxel size of its representation.
Even though \gls{nsvf}-small is \revised{\SI{2.5}{\decibel}} lower in quality compared to \ours-$2$, it takes $27\times$ longer to evaluate.
\gls{nsvf}-medium and \gls{nsvf}-large are similar in quality to \ours-$2$ and \ours-$4$ respectively, but take $43\times$ times longer to evaluate.
As \gls{nsvf} only uses a single \gls{mlp}, using fewer voxels can be advantageous if a very low memory footprint is required at the cost of quality, and \gls{nsvf}-small is the smallest scene representation in our comparison.
However, to achieve competitive quality, \gls{nsvf} requires more than $2\times$ the amount of memory of \ours.
Additionally, \gls{nsvf} takes about $6\times$ longer to train than \ours and also required ground truth depth for initialization for our challenging scenes.
Thus, for every configuration, a \ours exists that provides identical or better \emph{quality} at significantly better \emph{compactness} that can be evaluated much more \emph{efficiently}.

\paragraph*{\gls{llff}}
Compared to purely image-based methods like \gls{llff}, the advantages of \emph{neural} scene representations become apparent.
Even when using the entire training set as the basis for image-based rendering (and thus requiring more than $1000\times$ the memory of \ours) and cropping the image to remove border artifacts, \gls{llff} achieves the lowest quality among all tested methods.
Only for the highly challenging views of \forest and \sanmiguel, PSNR values are close to \gls{nerf} but still \revised{\SIrange{1}{7}{\decibel}} from \ours-2.
However, due to its simplicity of only selecting few, small light fields for each pose at test time and blending between them, \gls{llff} is the fastest method for novel view generation.
Thus, if memory is no concern, \gls{llff} may be the preferred option for low-power rendering.

\paragraph*{\gls{nex}}
\gls{nex} shows similar artifacts to \gls{llff}, in that it suffers from its explicit \gls{mpi} representation when generating views that differ too much from its reference view.
Although it is very fast to evaluate ($45\times$ faster than \ours) and achieves competitive quality for \sanmiguel, overall it requires $25\times$ more memory compared to \ours, and its quality is \SI{3}{\decibel} lower than \ours-2 on average.
Looking at the improved quality of \gls{nex}-\gls{mlp}, we can confirm that transferring its neural scene representation into an \gls{mpi} comes at a significant loss in quality.
Compared to \gls{nex}-\gls{mlp}, \ours achieves better quality using only $4-8$ samples for most scenes, at a $10\times$ speedup and $25\times$ more compactness.
For simple scenes, limited fields of view, and if compactness is no concern, precomputing an \gls{mpi} from a neural scene representation enables high-resolution real-time performance.
However, when targeting higher quality or streaming, a complete neural scene representation, like \ours, is the better choice.
Finally, \gls{nerf}-like representations have also been shown to support dynamic scenes and relighting, which may tilt the scales further towards an approach like \ours, where the memory consumption of \gls{mpi} videos becomes even more prohibitive.

\begin{figure}
	\centering
	\captionsetup{labelfont=bf,textfont=it,justification=centering}
	\begin{subfigure}{.53\linewidth}
		\includegraphics[width=\linewidth]{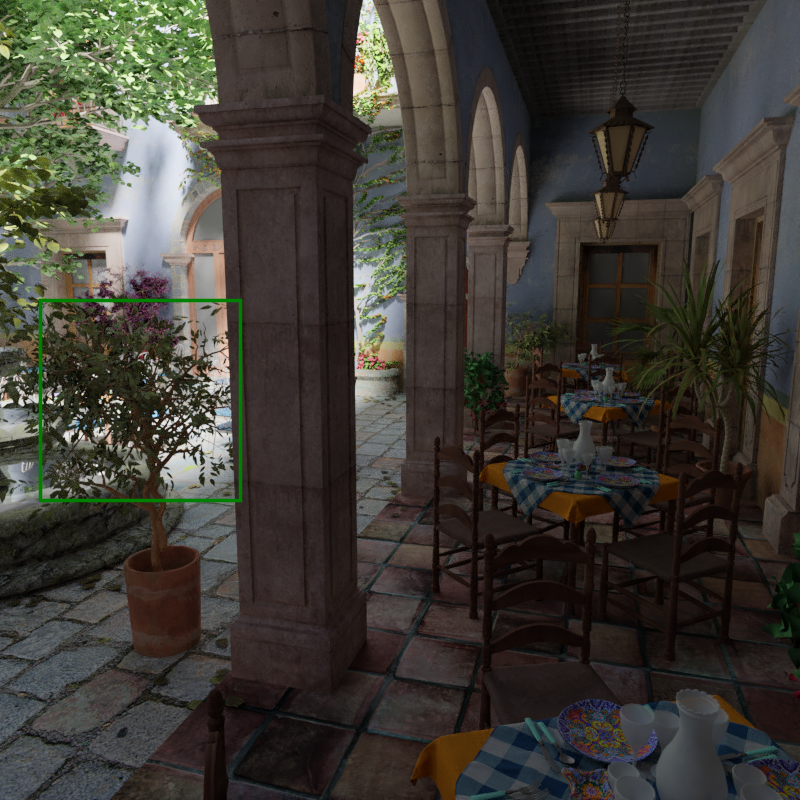}
		\caption{Ground Truth}
	\end{subfigure}%
	\hfill%
	\begin{subfigure}{.22\linewidth}
		\vspace{10pt}
		\includegraphics[width=\linewidth]{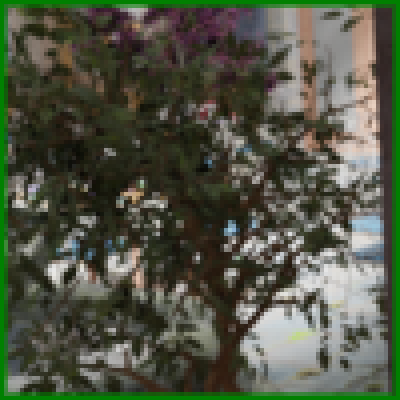}%
		\caption{GT}
		\centering
		\vspace{4.5pt}
		\includegraphics[width=1.0\linewidth]{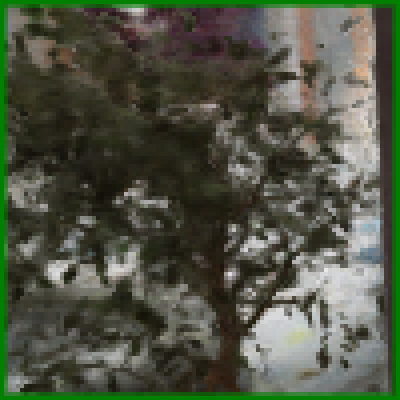}%
		\caption{\ours-4 \newline}
	\end{subfigure}%
	\hfill%
	\begin{subfigure}{.22\linewidth}
		\vspace{10pt}
		\includegraphics[width=\linewidth]{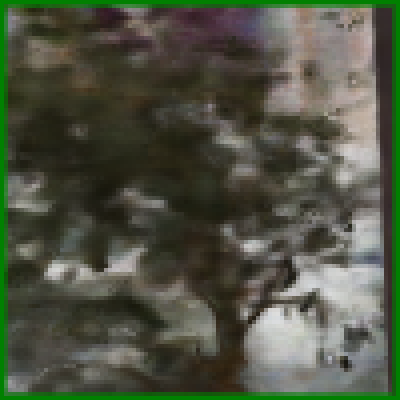}%
		\caption{\gls{nerf}}
		\centering
		\vspace{4.5pt}
		\includegraphics[width=1.0\linewidth]{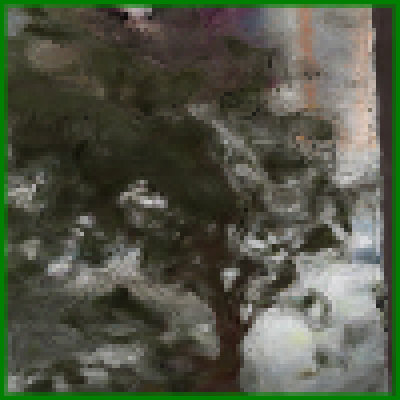}%
		\caption{\ours-4-noGT}
	\end{subfigure}%
	\captionsetup{labelfont=bf,textfont=it,justification=justified}
	\caption{\revised{(a) Example ground truth view of the test set of \sanmiguel with the corresponding inset (b). (c) \ours manages to preserve high-frequency detail around the leaves of the tree at only 4 samples per ray. (d) Although \gls{nerf} also preserves the tree faithfully, the edges are blurrier. (e) Even when trained with extracted \gls{nerf} depth maps (no GT depth), \ours  produces slightly sharper results than \gls{nerf}.}}
	\label{fig:sanmiguel_screenshot_comparison}
\end{figure}

\begin{figure}[t]
	\centering
	\captionsetup{labelfont=bf,textfont=it,justification=centering}
	\begin{subfigure}{.509\linewidth}
		\includegraphics[width=\linewidth]{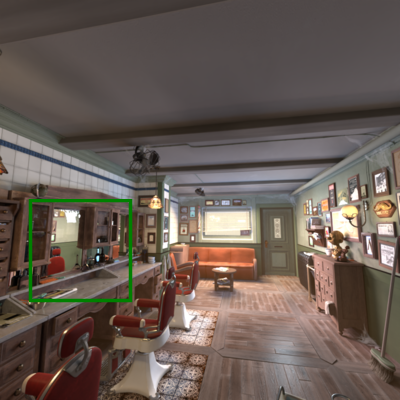}\\%
		\vspace{1pt}%
		\centering
		\includegraphics[width=\linewidth]{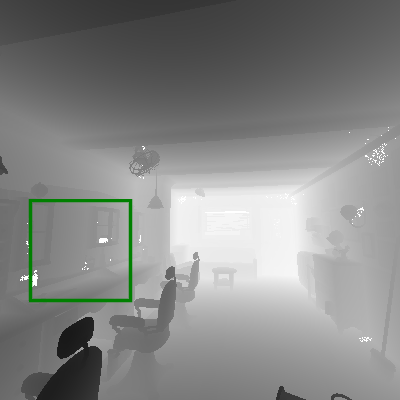}
		\caption{Ground Truth}
	\end{subfigure}%
	\hfill%
	\begin{subfigure}{.23\linewidth}
		\vspace{9.5pt}
		\includegraphics[width=\linewidth]{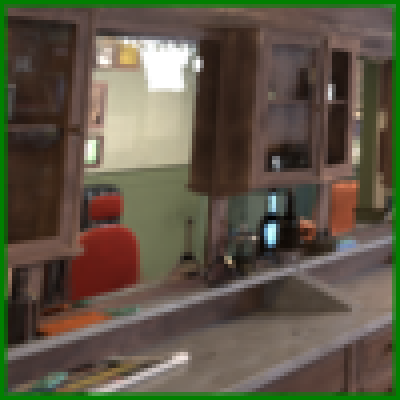}%
		\centering
		\vspace{1pt}
		\includegraphics[width=1.0\linewidth]{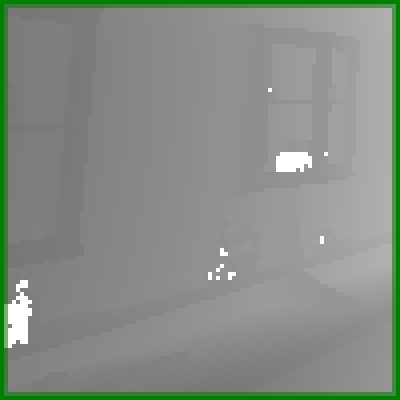}%
		\caption{GT}
		\vspace{3.7pt}
		\includegraphics[width=1.0\linewidth]{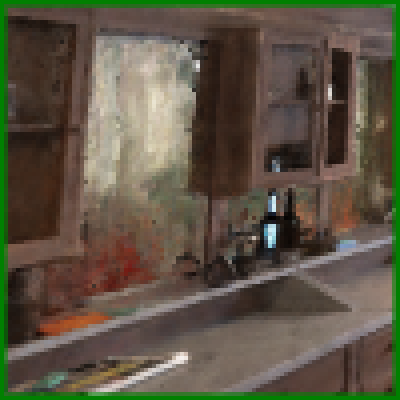}%
		\vspace{1pt}
		\includegraphics[width=1.0\linewidth]{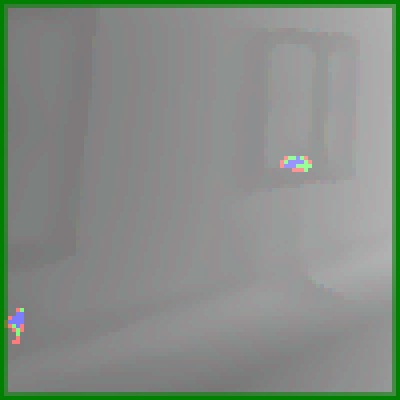}%
		\caption{\ours-4 \newline}
	\end{subfigure}%
	\hfill%
	\begin{subfigure}{.23\linewidth}
		\vspace{9.5pt}
		\includegraphics[width=\linewidth]{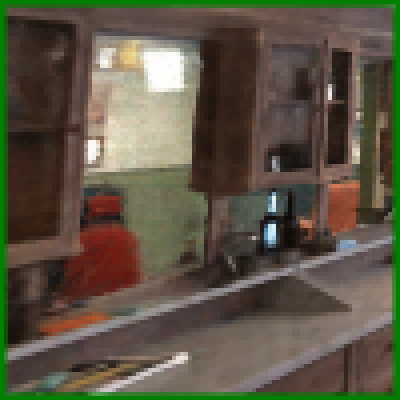}%
		\centering
		\vspace{1pt}
		\includegraphics[width=1.0\linewidth]{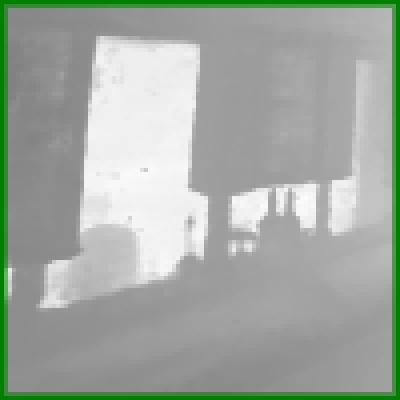}%
		\caption{\gls{nerf}}
		\vspace{3.7pt}
		\includegraphics[width=1.0\linewidth]{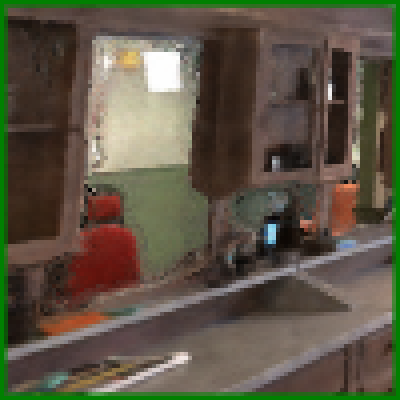}%
		\vspace{1pt}
		\includegraphics[width=1.0\linewidth]{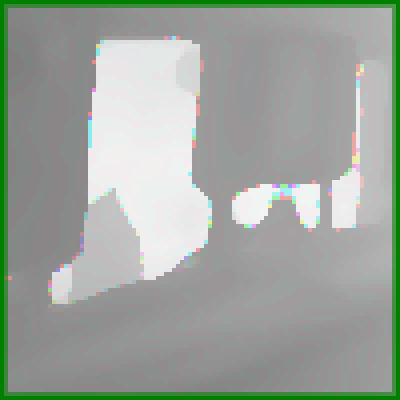}%
		\caption{\ours-4-noGT}
	\end{subfigure}%
	\captionsetup{labelfont=bf,textfont=it,justification=justified}
	\caption{\revised{(a) Example image and ground truth depth from the \barbershop test set with the corresponding inset (b). (c) The reflective mirrors cannot be accurately represented by just sampling around the mirror's depth in \ours. (d) \gls{nerf} places samples behind the mirror, constructing a virtual reflected room. (e) As depth maps extracted from \gls{nerf} include the virtual room, \ours-$4$-noGT learns to represent the mirrors with much fewer samples than \gls{nerf}.}}
	\label{fig:barbershop_screenshot_comparison}
\end{figure}

\subsection{DONeRF without Ground Truth Depth}%
Our proof-of-concept \ours that is trained without available ground truth depth (\ours-$X$-noGT) loses about \SI{1}{\decibel} on average compared to \ours (see Table~\ref{tbl:main_results}).
The losses are largest for \bulldozer; for \pavillon and \barbershop \ours-$X$-noGT actually outperforms \ours for higher sample counts.
For \bulldozer, \gls{nerf} outputs high frequency depth values across the entire background, which impedes the task of the depth oracle, only to create samples with zero contribution during shading---adding background detection to remove those samples during oracle training would likely resolve this issue.
Looking at \sanmiguel (Figure~\ref{fig:sanmiguel_screenshot_comparison}), \ours is capable of reconstructing fine details for foliage.
\ours-$X$-noGT interestingly produces sharper results than \gls{nerf} although we use a single \gls{nerf} depth output as ground truth depth.
We attribute \ours's ability to recover such high quality to filtering the depth targets and being able to place samples all around the foliage.

Even more surprising is the fact that \ours-$X$-noGT can outperform \ours, as in \pavillon and \barbershop (see Figure~\ref{fig:barbershop_screenshot_comparison}).
Transparent surfaces and fully reflective mirrors pose an issue for \ours, as the low sample count combined with only a single depth estimate is not sufficient for the network to perfectly replicate these challenging view-dependent phenomena.
However, for these scenes, \gls{nerf} essentially learns to place samples at multiple surfaces (for the refractive water) and in a virtual mirrored room (for the reflective mirror), and thus \ours-$X$-noGT places samples better than \ours for these parts and can reach higher quality for increased sample counts.

Overall, these results show that even without ground truth depth to train its depth oracle network, \ours is able to achieve better quality compared to \gls{nerf} at much lower sample counts, and provides the best tradeoff in terms of quality, performance and storage requirements.
Thus, a perfectly accurate depth estimate is not necessary to benefit from using \ours.

\section{Conclusion, Limitations and Future Work}

Starting from an evaluation of the sampling strategies of \gls{nerf}, we showed that sampling non-linearly and warped towards a view cell is beneficial in a variety of scenes.
Using ground truth depth for optimal sample placement, we showed that \emph{local sampling} achieves equal quality with as few as $2$ samples per ray compared to a fully sampled network with \num{128} samples. 
From this insight, we proposed our \emph{classified depth oracle network} which discretizes the space along each ray, and spatially filters the target across \emph{x}, \emph{y} and \emph{depth} to further improve the sample placement for challenging geometric scenarios. 
Using our oracle network to guide sample placement for a raymarched shading network, our compact \ours approach achieves equal or better quality compared to \gls{nerf} with 256 samples across most scenes, while using as few as \num{2} samples per ray at the memory requirement of only two \glspl{mlp}.
Compared to other scene representations and light field methods, \ours compares favorably in terms of storage requirements by a large margin and outperforms all other methods in quality.
Only image-based methods can be rendered faster than \ours.
Nevertheless, \ours makes a big step towards rendering directly from \emph{neural} scene representations with all their advantages in real-time.

Partially transparent objects 
and mirror-like surfaces
can pose an issue when using standard depth maps for training, as the first surface depth value does not represent the necessary sample locations along the ray. 
Fortunately, due to being a classification network, it would be straightforward to extend the training target of the depth oracle network with multiple ground truth surface points. %
The proof-of-concept \ours that is trained by using depth maps extracted from a \gls{nerf} already shows promising results for these surfaces. %
However, both, using multiple depth values for ground truth depth training, and eliminating the round trip through a full \gls{nerf} when there is no ground truth depth, are obvious next steps for \ours.

Furthermore, we only focused on static content in our evaluation.
For dynamic content, related research has already shown that depth-aware losses can be introduced to achieve more consistency~\cite{xian2020space, li2020neural}. %
Our classified depth sampling strategy could be adapted as a variation of these ideas, allowing for more consistency across dynamic content while staying within a compact neural representation.
Another partially orthogonal approach to ours is caching and baking \glspl{nerf} to further increase evaluation speeds.
Integrating an oracle, especially in combination with dynamic caching, may allow for further increases in rendering efficiency without compromising compactness for streaming.

To our knowledge, \ours is the first reliable method to render from a raymarched neural scene representation at interactive frame rates without exhaustive caching, and opens the door for compact high-quality dynamic rendering in real-time.
We are confident that such a local sampling strategy will be essential for real-time neural rendering going forward.

\printbibliography            

\appendix

\section{Additional Results: Efficient Neural Sampling}
\label{app:nerf_sampling}
Tables~\ref{tbl:bulldozer_ablation}, \ref{tbl:forest_ablation}, \ref{tbl:classroom_ablation} and \ref{tbl:sanmiguel_ablation} show per-scene results for varying numbers of samples per ray, using the sampling methods described in Section~\ref{sec:nerf_sampling}.

\section{Additional Results: Sampling Oracle Network}
\label{app:sampling_oracle}
Tables~\ref{tbl:bulldozer_filter}, \ref{tbl:forest_filter}, \ref{tbl:classroom_filter} and \ref{tbl:sanmiguel_filter} show per-scene results for varying numbers of samples per ray and various sampling oracle configurations. 
Section~\ref{sec:sampling_oracle_network} details the various depth oracle configurations.

\section{Additional Evaluation Setup: Datasets}
\label{app:scenes}
\bulldozer (by ''Heinzelnisse`` {\footnotesize \url{https://www.blendswap.com/blend/11490}}) [view cell size: $x=1, y=1, z=1$] shows a close view of a toy bulldozer made of building bricks with fine, high-frequency details and mostly diffuse shading. 
\revised{This scene differs from the dataset used in the original \gls{nerf}~\cite{mildenhall2020nerf}, as we re-rendered it to better fit our view cell methodology.} \\
\forest (by Robin Tran {\footnotesize \url{https://cloud.blender.org/p/gallery/5fbd186ec57d586577c57417}}) [$x=2, y=2, z=2$] shows a vast field of cel-shaded  high-frequency foliage and trees, with a person in the foreground, which enforces a good foreground and background representation. \\
\classroom (by Christophe Seux {\footnotesize \url{https://download.blender.org/demo/test/classroom.zip}}) [$x=0.7, y=0.7, z=0.2$] provides a challenging indoors scenario, with thin geometric detail such as the chairs' legs, as well as very detailed texture work. \\
\sanmiguel (by Guillermo M. Leal Llaguno {\footnotesize \url{https://casual-effects.com/g3d/data10/index.html}}) [$x=1, y=1, z=0.4$] provides a test on a proven computer graphics test scene with difficult high-frequency transparent materials (the leaves of the trees) as well as complex disocclusions. \\
\pavillon (by Hamza Cheggour / ''eMirage`` {\footnotesize \url{https://download.blender.org/demo/test/pabellon_barcelona_v1.scene_.zip}}) [ $x=1, y=1, z=1$] contains complex view-dependent effects such as reflection and refraction in the pool as well as completely transparent objects. \\
\barbershop (by ''Blender Animation Studio`` {\footnotesize \url{https://svn.blender.org/svnroot/bf-blender/trunk/lib/benchmarks/cycles/barbershop_interior/}}) [$x=1.5, y=1.5, z=0.4$] is a small indoors scene containing realistic global illumination and challenging mirrors.

\section{Additional Evaluation Setup: \ours}
\label{app:ours}
\begin{figure}
	\centering
	\captionsetup{labelfont=bf,textfont=it}
	\begin{subfigure}[t]{.97\linewidth}
		\includegraphics[width=\linewidth]{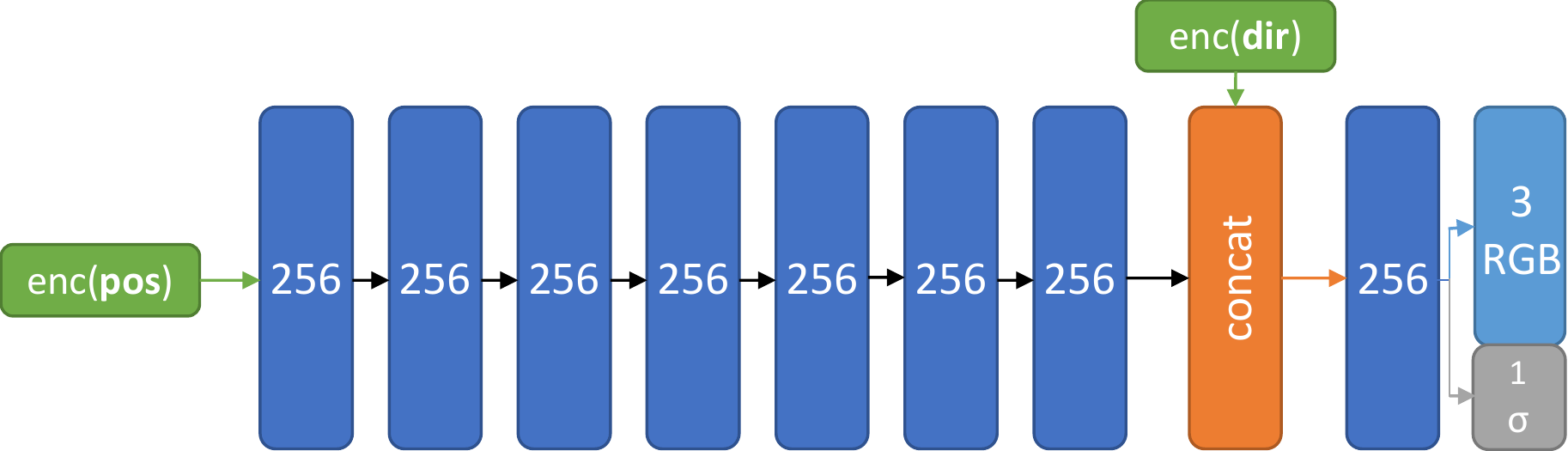}
	\end{subfigure}%
	
	\caption{For evaluation, we use the same \gls{mlp} architecture for both \gls{nerf} and \ours: $8$ layers with $256$ hidden units each, with a single skip connection to forward the encoded directions to the last layer. For the depth oracle network, we do not use a skip connection. The inputs and outputs are described in Section~\ref{sec:classified_depth}. }
	\label{fig:network_architecture}
\end{figure}

We use a slightly adapted version of the network architecture of \gls{nerf} for both networks of \ours and \gls{nerf} in our evaluation (Figure~\ref{fig:network_architecture}): slightly less capacity with $8$ layers at $256$ hidden units and only a single skip connection for the encoded directions. 
The depth oracle network neither uses skip connections nor encoding for the positions and directions, but still uses ray unification and non-linear sampling  (see Section~\ref{sec:sampling_oracle_network}).

\section{Additional Evaluation Setup: Baselines}
\label{app:baselines}
For each baseline method, we converted our datasets to use their required format, \ie, we provide ground truth poses and undistorted ground truth images for each method. 
We count the storage by summing all required network weights, checkpoints or images that are necessary to render novel views.
FLOP per pixel are counted either by profiling with NVIDIA Nsight Compute or directly from the neural network evaluation.

\paragraph*{\gls{nerf}~\cite{mildenhall2020nerf}}
We use an open source PyTorch port of \gls{nerf} ({\footnotesize \url{https://github.com/yenchenlin/nerf-pytorch}}) in our evaluation setup.
For consistency, we use the same network architecture as used for evaluating \ours (Figure~\ref{fig:network_architecture}) and use $64$ coarse and $128$ additional fine samples (resulting in $64+64+128=256$ total network evaluations) as in the original \gls{nerf}~\cite{mildenhall2020nerf}.
We train \gls{nerf} at \num{300000} iterations total, taking $2048$ samples from $2$ images per iteration.

\paragraph*{\gls{nsvf}~\cite{liu2020neural}}
For \gls{nsvf}, we use the authors' open source code ({\footnotesize \url{https://github.com/facebookresearch/NSVF}}) and perform a grid search over the initial voxel size.
As \gls{nsvf} varies greatly in terms of quality and performance based on the initial voxel size, we choose $3$ representative variants by choosing the best version within three scenarios:
\emph{\gls{nsvf}-large} is selected by choosing the best quality, ignoring performance and memory constraints, \emph{\gls{nsvf}-medium} is aimed at the best quality-performance tradeoff, and \emph{\gls{nsvf}-small} aims at the lowest memory requirements.
We train each \gls{nsvf} for \num{150000} iterations using $4096$ samples per iteration.

\paragraph*{\gls{llff}~\cite{mildenhall2019llff}}
For \gls{llff}, we use the authors' available code ({\footnotesize \url{https://github.com/Fyusion/LLFF}}), and evaluate the quality metrics by first removing a border of \SI{10}{\percent} from each side of the image, as suggested by the authors.

\paragraph*{\gls{nex}~\cite{Wizadwongsa2021NeX}}
For \gls{nex}, we use the authors' open source code ({\footnotesize \url{https://github.com/nex-mpi/nex-code/}}) and their suggested parameters that fit into the memory budget of $11$ GB GPU RAM, \ie, $256$ hidden units for their main \gls{mlp}, $6$ layers and $12$ sublayers for their \gls{mpi}. 
This also provides a more competitive comparison in terms of storage requirements.
We use the authors' real-time web viewer to render images for the evaluation, and train for \num{300000} iterations with $4096$ samples per iteration.
Furthermore, we include an additional baseline for \gls{nex}, where we do not precompute the radiance and neural basis functions into an \gls{mpi}, and instead query them from the \glspl{mlp} directly during inference, to show the potential upper limit in quality.
Note that this \gls{nex}-\gls{mlp} variant has slightly higher storage requirements than \gls{nex}, as it contains the base color \gls{mpi} in 32-bit floating point format.

\begin{table}[!htp]\centering
	\captionsetup{labelfont=bf,textfont=it}
	\caption{Ablation results for various depth oracle configurations for the \bulldozer scene. Please refer to Section~\ref{sec:sampling_oracle_network} of the main paper for a detailed explanation on these depth oracle configurations.}\label{tbl:bulldozer_filter}
	\small
	\begin{tabular}{>{\em}lc@{\rtbs}c@{\rtbs}c@{\rtbs}c@{\rtbs}c@{\rtbs}c@{\rtbs}c@{\rtbs}c@{\rtbs}c@{\rtbs}}\toprule
		\textbf{\bulldozer} &\multicolumn{4}{c}{PSNR $\uparrow$} &\multicolumn{4}{c}{FLIP  $\downarrow$} \\\cmidrule{1-9}
		Method &N = 2 &N = 4 &N = 8 &N = 16 &N = 2 &N = 4 &N = 8 &N = 16 \\\midrule
		SD &25.187 &25.977 &26.684 &27.347 &0.096 &0.083 &0.074 &0.067 \\
		SD unified &26.714 &27.242 &27.736 &28.196 &0.079 &0.071 &0.066 &0.061 \\
		K-1 Z-1 I-1 &26.936 &28.776 &30.476 &31.926 &0.074 &0.061 &0.052 &0.047 \\
		K-5 Z-1 I-1 &30.435 &33.241 &34.893 &36.019 &0.059 &0.046 &0.041 &0.038 \\
		K-5 Z-1 I-128 &\textbf{31.442} &33.840 &35.248 &36.203 &\textbf{0.053} &0.044 &0.040 &0.037 \\
		K-5 Z-5 I-128 &31.351 &\textbf{34.253} &\textbf{36.071} &37.119 &0.054 &\textbf{0.043} &\textbf{0.038} &\textbf{0.036} \\
		K-9 Z-1 I-1 &30.511 &33.506 &35.441 &36.561 &0.058 &0.046 &0.040 &0.037 \\
		K-9 Z-1 I-128 &30.837 &33.931 &35.645 &36.661 &0.056 &0.044 &0.039 &0.037 \\
		K-9 Z-9 I-128 &29.497 &32.654 &35.156 &\textbf{37.132} &0.064 &0.050 &0.042 &0.036 \\
		\bottomrule
	\end{tabular}
\end{table}

\begin{table}[!htp]\centering
	\captionsetup{labelfont=bf,textfont=it}
	\caption{Ablation results for various depth oracle configurations for the \forest scene. Please refer to Section~\ref{sec:sampling_oracle_network} of the main paper for a detailed explanation on these depth oracle configurations.}\label{tbl:forest_filter}
	\small
	\begin{tabular}{>{\em}lc@{\rtbs}c@{\rtbs}c@{\rtbs}c@{\rtbs}c@{\rtbs}c@{\rtbs}c@{\rtbs}c@{\rtbs}c@{\rtbs}}\toprule
		\textbf{\forest} &\multicolumn{4}{c}{PSNR $\uparrow$} &\multicolumn{4}{c}{FLIP  $\downarrow$} \\\cmidrule{1-9}
		Method &N = 2 &N = 4 &N = 8 &N = 16 &N = 2 &N = 4 &N = 8 &N = 16 \\\midrule
		SD &29.884 &30.640 &31.832 &33.212 &0.067 &0.062 &0.057 &0.053 \\
		SD unified &30.768 &31.608 &32.822 &33.876 &0.063 &0.058 &0.054 &0.050 \\
		K-1 Z-1 I-1 &29.261 &30.933 &32.425 &33.501 &0.069 &0.061 &0.056 &0.052 \\
		K-5 Z-1 I-1 &30.262 &32.535 &34.057 &34.978 &0.064 &0.056 &0.051 &0.050 \\
		K-5 Z-1 I-128 &32.188 &\textbf{34.327} &35.119 &35.591 &\textbf{0.056} &\textbf{0.051} &0.049 &0.048 \\
		K-5 Z-5 I-128 &\textbf{32.395} &34.177 &\textbf{35.385} &\textbf{36.308} &\textbf{0.056} &\textbf{0.051} &\textbf{0.047} &\textbf{0.045} \\
		K-9 Z-1 I-1 &30.617 &33.203 &34.920 &35.648 &0.063 &0.054 &0.049 &0.048 \\
		K-9 Z-1 I-128 &31.136 &33.861 &35.298 &35.869 &0.061 &0.052 &0.049 &0.047 \\
		K-9 Z-9 I-128 &30.558 &32.941 &34.452 &35.830 &0.064 &0.054 &0.050 &0.046 \\
		\bottomrule
	\end{tabular}
\end{table}

\begin{table}[!htp]\centering
	\captionsetup{labelfont=bf,textfont=it}
	\caption{Ablation results for various depth oracle configurations for the \classroom scene. Please refer to Section~\ref{sec:sampling_oracle_network} of the main paper for a detailed explanation on these depth oracle configurations.}\label{tbl:classroom_filter}
	\small
	\begin{tabular}{>{\em}lc@{\rtbs}c@{\rtbs}c@{\rtbs}c@{\rtbs}c@{\rtbs}c@{\rtbs}c@{\rtbs}c@{\rtbs}c@{\rtbs}}\toprule
		\textbf{\classroom} &\multicolumn{4}{c}{PSNR $\uparrow$} &\multicolumn{4}{c}{FLIP  $\downarrow$} \\\cmidrule{1-9}
		Method &N = 2 &N = 4 &N = 8 &N = 16 &N = 2 &N = 4 &N = 8 &N = 16 \\\midrule
		SD &26.832 &27.403 &28.205 &29.601 &0.103 &0.095 &0.086 &0.076 \\
		SD unified &27.732 &28.315 &29.233 &30.351 &0.091 &0.086 &0.078 &0.071 \\
		K-1 Z-1 I-1 &27.449 &28.882 &29.587 &31.733 &0.092 &0.081 &0.079 &0.068 \\
		K-5 Z-1 I-1 &28.697 &30.506 &30.823 &32.998 &0.083 &0.072 &0.074 &0.063 \\
		K-5 Z-1 I-128 &29.965 &31.882 &33.236 &33.581 &0.073 &0.069 &0.061 &0.061 \\
		K-5 Z-5 I-128 &\textbf{30.749} &\textbf{32.244} &33.482 &\textbf{34.348} &\textbf{0.070} &0.067 &0.060 &\textbf{0.058} \\
		K-9 Z-1 I-1 &29.387 &31.640 &32.681 &33.836 &0.078 &\textbf{0.067} &0.066 &0.059 \\
		K-9 Z-1 I-128 &30.006 &32.200 &\textbf{33.619} &33.984 &0.073 &0.068 &\textbf{0.058} &0.059 \\
		K-9 Z-9 I-128 &29.698 &31.574 &32.656 &34.029 &0.081 &0.069 &0.062 &0.059 \\
		\bottomrule
	\end{tabular}
\end{table}

\begin{table}[!htp]\centering
	\captionsetup{labelfont=bf,textfont=it}
	\caption{Ablation results for various depth oracle configurations for the \sanmiguel scene. Please refer to Section~\ref{sec:sampling_oracle_network} of the main paper for a detailed explanation on these depth oracle configurations.}\label{tbl:sanmiguel_filter}
	\small
	\begin{tabular}{>{\em}lc@{\rtbs}c@{\rtbs}c@{\rtbs}c@{\rtbs}c@{\rtbs}c@{\rtbs}c@{\rtbs}c@{\rtbs}c@{\rtbs}}\toprule
		\textbf{\sanmiguel} &\multicolumn{4}{c}{PSNR $\uparrow$} &\multicolumn{4}{c}{FLIP  $\downarrow$} \\\cmidrule{1-9}
		Method &N = 2 &N = 4 &N = 8 &N = 16 &N = 2 &N = 4 &N = 8 &N = 16 \\\midrule
		SD &25.314 &25.699 &26.283 &27.163 &0.097 &0.092 &0.086 &0.081 \\
		SD unified &25.142 &25.487 &26.078 &26.925 &0.099 &0.094 &0.088 &0.082 \\
		K-1 Z-1 I-1 &26.080 &26.977 &27.970 &28.746 &0.092 &0.082 &0.075 &0.070 \\
		K-5 Z-1 I-1 &26.406 &27.580 &28.547 &29.364 &0.086 &0.078 &0.071 &0.066 \\
		K-5 Z-1 I-128 &\textbf{27.125} &27.986 &28.992 &29.739 &\textbf{0.079} &\textbf{0.073} &0.068 &\textbf{0.064} \\
		K-5 Z-5 I-128 &27.105 &27.980 &28.906 &29.781 &\textbf{0.079} &0.074 &0.068 &\textbf{0.064} \\
		K-9 Z-1 I-1 &26.505 &27.808 &29.056 &29.815 &0.086 &0.076 &0.068 &0.065 \\
		K-9 Z-1 I-128 &26.875 &\textbf{28.200} &\textbf{29.257} &\textbf{29.893} &0.082 &0.073 &\textbf{0.066} &\textbf{0.064} \\
		K-9 Z-9 I-128 &26.330 &27.571 &28.484 &29.431 &0.088 &0.078 &0.071 &0.066 \\
		\bottomrule
	\end{tabular}
\end{table}

\begin{table*}[!htp]\centering
	\captionsetup{labelfont=bf,textfont=it}
	\caption{Ablation results for various sampling methods for the \bulldozer scene. Please refer to Section~\ref{sec:nerf_sampling} of the main paper for a detailed explanation on these sampling strategies.}\label{tbl:bulldozer_ablation}
	\small
	\begin{tabular}{lcc@{\rtbsx}cc@{\rtbsx}cc@{\rtbsx}cc@{\rtbsx}cc@{\rtbsx}cc@{\rtbsx}cc@{\rtbsx}cc@{\rtbsx}cc}\toprule
		\textbf{\bulldozer} & &\multicolumn{2}{c}{uniform} &\multicolumn{2}{c}{logarithmic} &\multicolumn{2}{c}{log+warp} &\multicolumn{2}{c}{NDC} &\multicolumn{2}{c}{\shortstack{uniform\\ local}} &\multicolumn{2}{c}{\shortstack{logarithmic\\ local}} &\multicolumn{2}{c}{\shortstack{log+warp\\ local}} &\multicolumn{2}{c}{\shortstack{NDC\\ local}} \\\midrule
		& &PSNR &FLIP &PSNR &FLIP &PSNR &FLIP &PSNR &FLIP &PSNR &FLIP &PSNR &FLIP &PSNR &FLIP &PSNR &FLIP \\\midrule
		\multirow{7}{*}{\shortstack[l]{Number of \\ samples N}} &2 & & & & & & & & &27.930 &0.062 &27.774 &0.063 &27.755 &0.064 &25.158 &0.099 \\
		&4 &16.662 &0.276 &16.562 &0.276 &16.834 &0.275 &11.929 &0.408 &28.071 &0.061 &28.038 &0.060 &28.034 &0.060 &26.147 &0.086 \\
		&8 &19.720 &0.204 &19.458 &0.209 &19.637 &0.206 &14.814 &0.314 &28.115 &0.061 &28.093 &0.059 &28.129 &0.059 &26.358 &0.083 \\
		&16 &22.790 &0.146 &22.683 &0.147 &22.769 &0.145 &18.384 &0.229 &28.152 &0.061 &28.169 &0.059 &28.192 &0.059 &26.483 &0.084 \\
		&32 &26.551 &0.096 &26.233 &0.098 &26.458 &0.095 &22.027 &0.156 &28.208 &0.060 &28.223 &0.058 &28.269 &0.059 &26.891 &0.081 \\
		&64 &30.660 &0.064 &30.472 &0.064 &30.817 &0.062 &25.381 &0.106 &28.366 &0.060 &28.939 &0.055 &28.975 &0.056 &29.424 &0.068 \\
		&128 &34.362 &0.047 &34.261 &0.047 &34.651 &0.045 &29.265 &0.075 &30.169 &0.052 &32.551 &0.047 &32.640 &0.047 &29.172 &0.071 \\
		\bottomrule
	\end{tabular}
\end{table*}

\begin{table*}[!htp]\centering
	\captionsetup{labelfont=bf,textfont=it}
	\caption{Ablation results for various sampling methods for the \forest scene. Please refer to Section~\ref{sec:nerf_sampling} of the main paper for a detailed explanation on these sampling strategies.}\label{tbl:forest_ablation}
	\small
	\begin{tabular}{lcc@{\rtbsx}cc@{\rtbsx}cc@{\rtbsx}cc@{\rtbsx}cc@{\rtbsx}cc@{\rtbsx}cc@{\rtbsx}cc@{\rtbsx}cc}\toprule
		\textbf{\forest} & &\multicolumn{2}{c}{uniform} &\multicolumn{2}{c}{logarithmic} &\multicolumn{2}{c}{log+warp} &\multicolumn{2}{c}{NDC} &\multicolumn{2}{c}{\shortstack{uniform\\ local}} &\multicolumn{2}{c}{\shortstack{logarithmic\\ local}} &\multicolumn{2}{c}{\shortstack{log+warp\\ local}} &\multicolumn{2}{c}{\shortstack{NDC\\ local}} \\\midrule
		\multirow{7}{*}{\shortstack[l]{Number of \\ samples N}} &2 & & & & & & & & &24.126 &0.156 &29.666 &0.079 &30.867 &0.071 &22.108 &0.182 \\
		&4 &14.794 &0.525 &21.062 &0.227 &21.185 &0.220 &15.431 &0.474 &27.384 &0.102 &28.513 &0.089 &30.760 &0.075 &22.683 &0.170 \\
		&8 &20.172 &0.242 &22.494 &0.171 &23.129 &0.160 &18.651 &0.305 &27.164 &0.104 &27.909 &0.100 &29.799 &0.087 &22.801 &0.169 \\
		&16 &20.282 &0.239 &24.430 &0.144 &25.363 &0.131 &21.185 &0.221 &26.724 &0.109 &26.762 &0.117 &29.070 &0.091 &23.417 &0.164 \\
		&32 &20.824 &0.229 &25.294 &0.139 &25.995 &0.129 &23.403 &0.166 &22.786 &0.179 &25.492 &0.133 &27.754 &0.104 &23.581 &0.164 \\
		&64 &22.498 &0.178 &27.357 &0.113 &26.951 &0.118 &25.962 &0.122 &22.137 &0.194 &24.996 &0.150 &26.242 &0.129 &22.961 &0.170 \\
		&128 &23.082 &0.167 &26.536 &0.125 &28.809 &0.100 &29.291 &0.097 &22.416 &0.184 &27.067 &0.123 &29.552 &0.097 &23.696 &0.164 \\
		\bottomrule
	\end{tabular}
\end{table*}

\begin{table*}[!htp]\centering
	\captionsetup{labelfont=bf,textfont=it}
	\caption{Ablation results for various sampling methods for the \classroom scene. Please refer to Section~\ref{sec:nerf_sampling} of the main paper for a detailed explanation on these sampling strategies.}\label{tbl:classroom_ablation}
	\small
	\begin{tabular}{lcc@{\rtbsx}cc@{\rtbsx}cc@{\rtbsx}cc@{\rtbsx}cc@{\rtbsx}cc@{\rtbsx}cc@{\rtbsx}cc@{\rtbsx}cc}\toprule
		\textbf{\classroom} & &\multicolumn{2}{c}{uniform} &\multicolumn{2}{c}{logarithmic} &\multicolumn{2}{c}{log+warp} &\multicolumn{2}{c}{NDC} &\multicolumn{2}{c}{\shortstack{uniform\\ local}} &\multicolumn{2}{c}{\shortstack{logarithmic\\ local}} &\multicolumn{2}{c}{\shortstack{log+warp\\ local}} &\multicolumn{2}{c}{\shortstack{NDC\\ local}} \\\midrule
		\multirow{7}{*}{\shortstack[l]{Number of \\ samples N}} &2 & & & & & & & & &33.389 &0.054 &33.668 &0.054 &33.458 &0.055 &32.157 &0.060 \\
		&4 &21.135 &0.194 &21.804 &0.204 &21.497 &0.211 &18.920 &0.275 &33.994 &0.052 &33.964 &0.053 &33.815 &0.054 &33.033 &0.057 \\
		&8 &24.357 &0.141 &25.263 &0.130 &25.689 &0.120 &22.690 &0.180 &33.889 &0.053 &33.949 &0.054 &33.948 &0.054 &33.259 &0.056 \\
		&16 &26.519 &0.111 &27.313 &0.105 &27.838 &0.094 &25.689 &0.123 &33.346 &0.055 &33.688 &0.054 &33.952 &0.053 &33.325 &0.056 \\
		&32 &28.882 &0.085 &29.678 &0.082 &30.435 &0.072 &28.342 &0.088 &33.495 &0.056 &33.786 &0.055 &33.883 &0.054 &33.587 &0.057 \\
		&64 &31.322 &0.068 &32.252 &0.064 &32.822 &0.060 &31.307 &0.065 &33.660 &0.057 &34.182 &0.055 &34.452 &0.053 &33.586 &0.058 \\
		&128 &33.113 &0.059 &33.781 &0.057 &34.527 &0.053 &33.823 &0.055 &33.574 &0.058 &33.727 &0.057 &34.510 &0.055 &33.674 &0.058 \\
		\bottomrule
	\end{tabular}
\end{table*}

\begin{table*}[!htp]\centering
	\captionsetup{labelfont=bf,textfont=it}
	\caption{Ablation results for various sampling methods for the \sanmiguel scene. Please refer to Section~\ref{sec:nerf_sampling} of the main paper for a detailed explanation on these sampling strategies.}\label{tbl:sanmiguel_ablation}
	\small
		\begin{tabular}{lcc@{\rtbsx}cc@{\rtbsx}cc@{\rtbsx}cc@{\rtbsx}cc@{\rtbsx}cc@{\rtbsx}cc@{\rtbsx}cc@{\rtbsx}cc}\toprule
			\textbf{\sanmiguel} & &\multicolumn{2}{c}{uniform} &\multicolumn{2}{c}{logarithmic} &\multicolumn{2}{c}{log+warp} &\multicolumn{2}{c}{NDC} &\multicolumn{2}{c}{\shortstack{uniform\\ local}} &\multicolumn{2}{c}{\shortstack{logarithmic\\ local}} &\multicolumn{2}{c}{\shortstack{log+warp\\ local}} &\multicolumn{2}{c}{\shortstack{NDC\\ local}} \\\midrule
		\multirow{7}{*}{\shortstack[l]{Number of \\ samples N}} &2 & & & & & & & & &28.689 &0.072 &28.658 &0.071 &28.919 &0.068 &28.286 &0.073 \\
		&4 &20.763 &0.213 &21.442 &0.197 &21.662 &0.190 &17.843 &0.291 &28.824 &0.071 &28.705 &0.071 &28.861 &0.070 &28.627 &0.072 \\
		&8 &22.607 &0.171 &23.610 &0.146 &24.031 &0.137 &21.549 &0.190 &28.810 &0.072 &28.632 &0.072 &28.879 &0.072 &28.913 &0.071 \\
		&16 &23.634 &0.149 &24.588 &0.128 &25.250 &0.114 &23.711 &0.138 &28.675 &0.075 &28.595 &0.074 &28.935 &0.072 &29.096 &0.071 \\
		&32 &24.795 &0.126 &25.369 &0.114 &26.429 &0.098 &25.702 &0.103 &28.654 &0.076 &28.533 &0.077 &29.027 &0.072 &29.330 &0.071 \\
		&64 &26.137 &0.105 &26.283 &0.102 &27.607 &0.086 &27.783 &0.080 &28.476 &0.079 &28.460 &0.079 &28.976 &0.074 &29.566 &0.070 \\
		&128 &27.227 &0.092 &27.391 &0.089 &28.825 &0.075 &29.453 &0.069 &28.498 &0.079 &28.132 &0.079 &28.869 &0.075 &29.582 &0.069 \\
		\bottomrule
	\end{tabular}
\end{table*}

\end{document}